\documentclass[final]{article}

\usepackage{arxiv}
\usepackage[utf8]{inputenc} 
\usepackage[T1]{fontenc}    
\usepackage{hyperref}       
\usepackage{url}            
\usepackage{booktabs}       
\usepackage{amsfonts}       
\usepackage{nicefrac}       
\usepackage{microtype}      
\usepackage{lipsum}
\usepackage{graphicx}

\usepackage{amssymb}
\usepackage{latexsym}
\usepackage{amsmath}
\usepackage{amsfonts}
\usepackage{mathrsfs}

\usepackage{algorithm}
\usepackage{algorithmic}

\usepackage{graphicx}
\usepackage{afterpage}
\usepackage{wrapfig}
\usepackage{bbm}
\usepackage{array}
\usepackage{booktabs}
\usepackage{caption}
\usepackage[table]{xcolor}
\usepackage{url}
\usepackage{hyperref}
\usepackage{lipsum}
\usepackage{CJKutf8}
\usepackage{multirow}
\usepackage{multicol}

\usepackage{lineno} 

\usepackage{float}
\usepackage{pdfpages}
\usepackage{changes}  
\usepackage{xcolor}
\definechangesauthor[name={Author1}, color=red]{A1} 
\definechangesauthor[name={Author2}, color=green]{A2} 

\definecolor{newcolor}{rgb}{.8,.349,.1}

\newtheorem{theorem}{Theorem}
\newtheorem{lemma}[theorem]{Lemma}
\newtheorem{proposition}{Proposition}

\title{Embedded Visual Prompt Tuning}

\author{Wenqiang Zu$^{1,4,6*}$, 
Shenghao Xie$^{2,5,6*}$, Qing Zhao$^{3}$\thanks{Equal contribution.},
 Guoqi Li$^{4}$ \\ \textbf{Lei Ma}$^{3,6}$\thanks{Lei Ma is the corresponding author with email: \textit{lei.ma@pku.edu.cn}.} \\
$^1$ School of Artificial Intelligence, University of Chinese Academy of Sciences \\
$^2$ Academy for Advanced Interdisciplinary Studies, Peking University \\
$^3$ College of Future Technology, Peking University\\
$^4$ Institute of Automation, Chinese Academy of Sciences\\
$^5$ School of Cyber Science and Engineering, Wuhan University\\
$^6$ Beijing Academy of Artificial Intelligence\\
}

\begin{document}
\maketitle

\newcommand{\xsh}[1]{\textcolor{black}{#1}}

\newcommand{\wq}[1]{\textcolor{black}{#1}}
\newcommand{\zdiff}[1]{\textcolor{black}{#1}}
\newcommand{\zdiffminor}[1]{\textcolor{black}{#1}}

\begin{abstract}
\textbf{Parameter-efficient fine-tuning (PEFT)} methods aim to adapt large scale pre-trained models to new domains by updating a small portion of parameters to reduce computational overhead. However, the effectiveness of these PEFT methods, especially in cross-domain few-shot scenarios, e.g., medical images, has not been fully explored. This paper facilitate the study of the performance of PEFT when fine-tuning pre-trained models on medical images. Furthermore, to alleviate the limitations of prompt introducing ways and approximation capabilities on Transformer architectures of mainstream prompt tuning methods, we propose the \textbf{Embedded Prompt Tuning (EPT)} method by embedding prompt tokens into the expanded channels. We also find that there are anomalies in the feature space distribution of pre-trained models during pre-training process, asnd prompt tuning can help mitigate this negative impact. To explain this phenomenon, we also introduce a novel perspective to understand prompt tuning: \textbf{Prompt tuning is a distribution calibrator.} And we support it by analyzing patch-wise scaling and feature separation operations contained in EPT. Our experiments show that EPT outperforms \zdiffminor{several state-of-the-art fine-tuning} methods by a significant margin on \zdiffminor{few-shot medical image classification}, and completes the fine-tuning process within highly competitive time, indicating EPT is an effective PEFT method. The source code is available at
\href{https://github.com/zuwenqiang/EPT}{github.com/zuwenqiang/EPT}.
\end{abstract}
\section{Introduction}
\xsh{Benifiting from massive training data, large scale pre-trained models~\cite{bommasani2021opportunities} have been widely witnessed to achieve impressive performance in various natural imaging downstream tasks, e.g., classification~\cite{zhang2023text, zhu2023pointclip}, segmentation~\cite{kirillov2023segment, cheng2023sam, wang2023sam, du2023segvol}, and detection~\cite{zhou2023foundation, chowdhury2023can, zhang2024fm}.} 

\xsh{Recent studies have attempted to explore the potential of pre-trained models \zdiffminor{in }medical image analysis\zdiffminor{~\cite{mazurowski2023segment,silva2023transductive,zhang2023customized,ma2024segment,dutt2023fairtune}}. Although pre-trained models have robust representation and generalization capabilities on natural images, their adaptabilities are still challenged when facing downstream tasks with a significant domain gap~\cite{wu2023can}. Therefore, pre-trained models need to be re-trained on \zdiffminor{medical} datasets to incorporate \zdiffminor{domain-specific} knowledge.}

\xsh{However, due to the huge number of parameters in pre-trained models, training from scratch would result in significant computational and memory cost. To address this issue, parameter-efficient fine-tuning (PEFT)~\cite{ding2023parameter} methods have been proposed. The main idea of PEFT is to freeze most of parameters in pre-trained models, and only fine-tune a small number of existing or additionally introduced parameters. \zdiffminor{In} medical image analysis, \zdiffminor{the data avalilability is limited} due to several factors, e.g., significant imaging noise\zdiffminor{~\cite{sagheer2020review}}, high annotation cost\zdiffminor{~\cite{willemink2020preparing}}, privacy policies\zdiffminor{~\cite{kaissis2020secure}}, and rare diseases\zdiffminor{~\cite{chen2023dynamic}} etc. \zdiffminor{These factors exacerbate the data-hungry nature of the pre-trained models~\cite{dutt2024parameter}. Therefore, there is a more urgent need for research on fine-tuning methods in the field of medical image analysis.} To our best knowledge, the effectiveness of PEFT in such cross-domain few-shot scenarios has not yet been fully evaluated. Therefore, we launch the first comprehensive benchmark test of PEFT for \zdiffminor{few-shot} medical image classification tasks on MedFMC~\cite{wang2023medfmc}.}

\xsh{Prompt tuning is a common paradigm of PEFT by introducing extra prompt tokens into embedding tokens of the input image. VPT~\cite{jia2022visual} prepends prompts before images parallelly, while VP~\cite{bahng2022exploring} adds prompts into images. Intuitively, previous prompt tuning methods have flaws in the way of introducing prompts. VPT cannot adjust each original token individually and fine-grainedly, whereas VP significantly disrupts the information contained in original tokens. Moreover, Wang et al.~\cite{wang2024universality} points that prompt tuning has limited approximation capabilities on Transformer architectures, performing less effectively compared to LoRA~\cite{hu2021lora}. Therefore, we ask: \textit{Can we design a new prompt tuning method that effectively enhances the information of original tokens\deleted{but also breaks through the limitations of approximation capabilities}?} In this paper, we propose the \textbf{Embedded Prompt Tuning (EPT)} method to answer this question. By embedding prompt tokens into expanded channels, EPT can introduce useful context to optimize input tokens fine-grainedly when maintaining the original information.}

\xsh{Additionally, we observe that pre-trained models pre-trained on natural images, when directly applied to downstream medical image classification tasks, would result in an apparent distance between samples from the same class in the feature space. Furthermore, prompt tuning helps mitigate this negative impact\zdiffminor{~\cite{huang2024fpt,zheng2024exploring}}, and we wonder: \textit{Is prompt tuning fundamentally a type of distribution calibrators?} \zdiff{To figure it out, we further analyze two micro-operations of EPT: patch-wise scaling and feature separation, and provide preliminary theoretical analysis. In our findings, patch-wise scaling shorten the intra-class distribution distance by drawing closer the representation of common features among samples from the same class}, and feature separation enhances the representation \cite{zhang2023understanding} and decoupling capabilites of features by expanding the input dimension.}

\xsh{In summary, our work proposes a novel PEFT method for cross-domain few-shot learning and medical image analysis. We hope our methods and findings can inspire and facilitate more valuable works in the future. Our contributions can be summarized as follows:}
\begin{enumerate}
    \item \xsh{We propose a novel parameter-efficient prompt tuning method, namely EPT, which not only addresses shortcomings in prompt introducing ways of previous prompt tuning methods, but also exhibits stronger approximation capabilites on Transformer architectures.}
    \item  \xsh{We develop a new perspective to understand prompt tuning: Prompt is a distribution calibrator. We also support it by analyzing patch-wise scaling and feature separation operations in EPT intuitively and theoretically.}
    \item  \xsh{We launch the first comprehensive benchmark evaluation of PEFT for medical image classification tasks on MedFMC, offering further inspirations for future research.}
    \item \xsh{Extensive experiments demonstrate that EPT achieves superior performance in cross-domain few-shot scenarios, e.g., medical image analysis, outperforming \zdiffminor{several state-of-the-art fine-tuning} methods \zdiffminor{on} \zdiffminor{few-shot medical image classification tasks} by a large margin, and completes the fine-tuning process within highly competitive time.}
\end{enumerate}
\section{Related Work}

\subsection{\xsh{Parameter-Efficient Fine-Tuning}}
\xsh{The goal of PEFT is to achieve higher performance by fine-tuning as few parameters as possible during the adaptation of foundation models for downstream tasks. PEFT generally fine-tunes Transformer at three positions: input, backbone, and linear head. Since the \zdiffminor{linear} head is essential for downstream tasks, it typically needs to be fine-tuned. Therefore, PEFT pays more attention on the input and backbone, and has correspondingly raised two representative methods: prompt tuning and adapter tuning.}

\xsh{Prompt tuning introduces extra tokens into the input image and fine-tunes them. VPT~\cite{jia2022visual} prepends prompt tokens before embedding tokens parallelly. DVPT~\cite{he2023dvpt} employs cross-attention mechanism between prompts and features to capture distribution-specific knowledge. VP~\cite{bahng2022exploring} adds values of prompt tokens and embedding tokens across all channels. EVP~\cite{liu2023explicit} primarily focuses on high-frequency components of features. \zdiff{CoOp~\cite{zhou2022learning} utilizes learnable text to addressing the issue of semantic consistency across different sentences. CoCoOp~\cite{zhou2022conditional} introduces a Meta-net to generate prompts and enhance the generalizability. KgCoOp~\cite{yao2023visual} constructed a regularization term to minize the gap between generated and crafted prompts.}}

\xsh{Adapter tuning introduces carefully designed structures into the backbone and fine-tunes them. Adaptformer~\cite{chen2022adaptformer} is the first to introduce adapters to vision transformers by replacing MLP in transformer layers with AdaptMLP. \zdiff{Tip-Adapter~\cite{zhang2022tip} leverages knowledge from the few-shot dataset as keys and values and supports feature querying to update the output from CLIP encoders without training. TaskRes~\cite{yu2023task} decomposes pre-trained and task-specific knowledge on the textual classifier. CLIP-Adapter~\cite{gao2024clip} conducts residual-style feature blending between adapters and pre-trained backbones.} LoRA~\cite{hu2021lora} decomposes model weights through low-rank matrices. SCT~\cite{zhao2024sct} leverages task-specific knowledge from salient channels of the feature map.}

\xsh{Our proposed \zdiff{EPT belongs to the category of prompt tuning, we think that prompt tuning, compared to adapter tuning, can more directly influence the input by introducing extra context, thus having a stronger effect on distribution calibration}. It is also worth noting that the prompt tuning methods, unlike commonly used hard prompts in context learning, introduces parameter-learnable soft prompts. Different from VPT and VP, EPT pays more attention on expanded embedding channels. \zdiff{Unlike PEFT for multimodal foundation models, e.g., CoOp, CLIP-Adapter, Tip-Adapter, and TaskRes etc., EPT is designed for image-only foundation models. In cross-domain few-shot scenarios, e.g., medical image analysis, there are inadequate high-quality image-text pairs, the textual information of labels is insufficient to provide rich features, while the potential of images has not yet been fully exploited.}}

\subsection{\xsh{PEFT for Medical Image Analysis}}
\xsh{PEFT begins to be adopted when transferring foundational models to medical image analysis scenarios recently\cite{dutt2024parameter}. DVPT~\cite{he2023dvpt}, as a variant of VPT, explores the potential of prompt tuning in medical image analysis scenarios. Zhang et al. adapts SAM~\cite{kirillov2023segment} to medical domains by LoRA. Wu et al.~\cite{wu2023medical} proposes medical sam adapter (Med-SA) to incorporate medical knowledge into SAM and designs Hyper-Prompting Adapter (HyP-Adpt) to adapt to different prompt conditioning strategies. However, the effectiveness of PEFT has yet to be fully evaluated in cross-domain few-shot scenarios, e.g., medical image analysis. VPPT~\cite{song2023vppt} conducts experiments of prompt tuning in natural imaging few-shot scenarios, but adapter tuning has not been tested and the domain gap is not significant. Dutt et al.~\cite{dutt2024parameter} tests the performance of adapter tuning in medical image analysis scenarios, but prompt tuning methods do not receive sufficient attention. Wang et al.~\cite{wang2023medfmc} proposed MedFMC benchmark in order to promote the research of PEFT during the adaptation of foundation models to medical domains. In this paper, we conduct the first comprehensive effectiveness evaluation of PEFT in cross-domain few-shot scenarios, e.g., medical image analysis.}

\begin{figure*}[!t]
  \centering
  \includegraphics[width=1\textwidth]{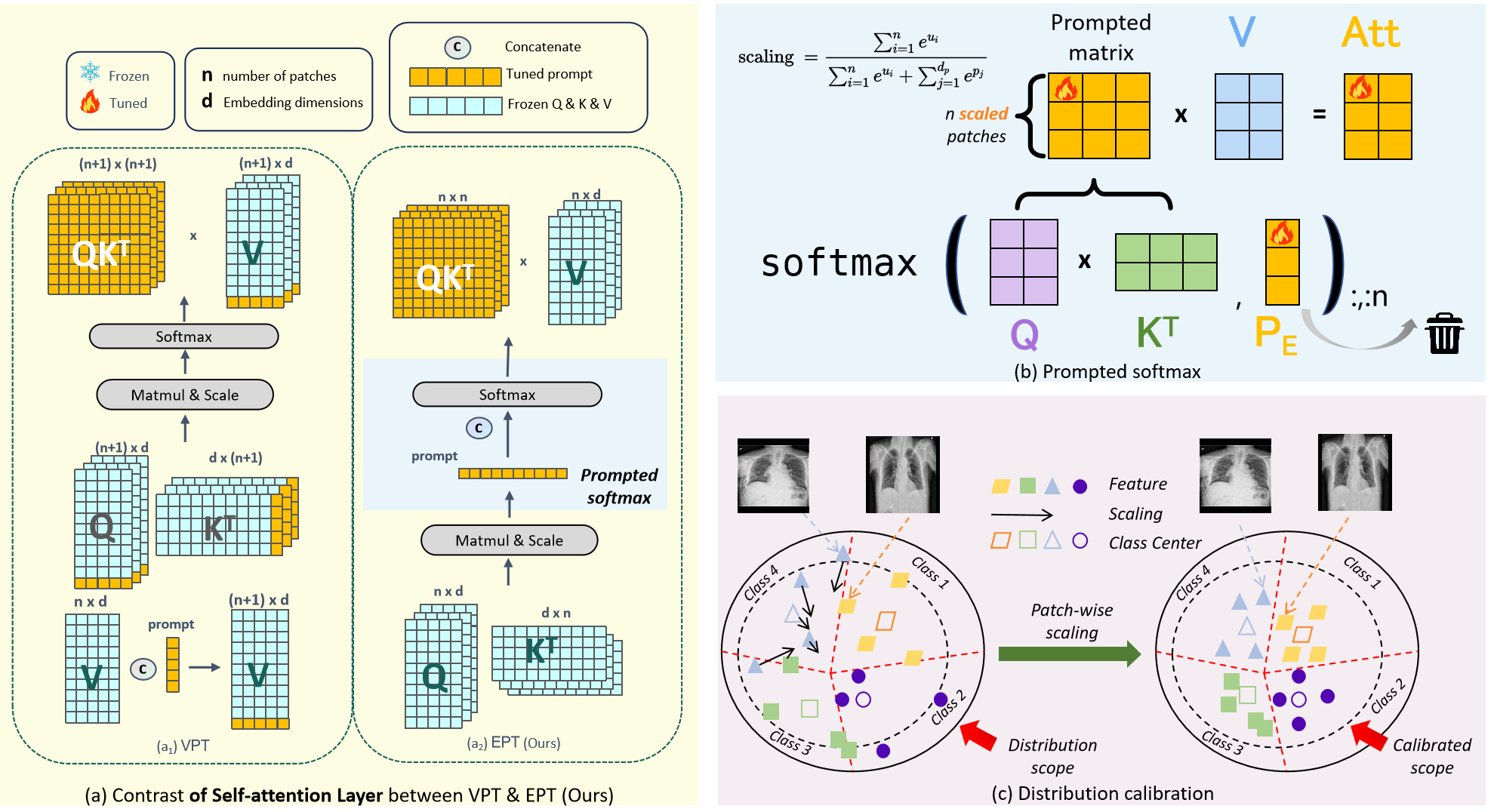}
  \caption{Overview of proposed \textbf{Embedded Prompt Tuning (EPT)}\deleted{and \textbf{Location-Aware Prompt (LAP)} Framework}. (a) \wq{Contrast of }VPT and EPT framework: During the fine-tuning process of \wq{VPT}\deleted{pre-trained vision Transformer}, only the prompts embedded in each input image token are updated while the whole Transformer encoder is frozen. As a comparison, \wq{EPT embeds prompts in the softmax operation of Q and K matrices in each token (patch), rather than directly embedding prompts in input positions}\deleted{EPT embeds each prompt token with the corresponding image tokens in the direction of embedding dimmensions and VPT prepends prompt tokens before image tokens in the direction of the width and height of the image}.  (b) \wq{Overview of prompted softmax. EPT achieves scaling of embedding for each patch by embedding (or concatenating after or before each patch) learnable parameters in the multiplication matrix of Q and K, followed by utilizing softmax normalization, thereby achieving scaling functionality for each patch embedding} \deleted{Contrast of Self-attention Layer between VPT and EPT (Ours): EPT adds prompts with the V matrix and concurrently concatenating prompts with the Q and K matrices in the direction of embedding dimmensions. In contrast, VPT inserts prompts in the direction of width and height to Q, K and V matrices, giving priority to increasing the input breadth instead of depth considered by EPT which may enhance the representation capability of input tokens}. (c) \wq{EPT achieves distribution calibration of frozen features through patch-wise scaling. Patch-wise scaling can reduce the intra-class distances for each class of samples.}\deleted{Location-Aware Prompt: For medical images, we first generate saliency maps to identify salient regions. We then select the top $K$ tokens, calculate the local prior of each selected token and the attention prior of all selected tokens respectively. At last, we add local priors with embeddings of corresponding each selected token and the classification token in prompts. (Here we generate the saliency map using the traditional machine learning based method from SSiT \cite{huang2022ssit} for better illustration.)}}
  \label{fig:figure1}
\end{figure*}

\section{Method}
\subsection{Preliminaries}
\paragraph{Notations and Organization}
Let $\sigma$ be the softmax operators. A bold lower case character, \textit{e.g.} $\mathbf{x}$, denotes a vector. A bold upper case character, \textit{e.g.} $\mathbf{W}$, denotes a matrix while $\mathbf{W}_{i, j}$, $\mathbf{W}_{i, { : }}$ and $\mathbf{W}_{:, j}$ is the $(i, j)$-th element, $i$-th row, $j$-th column, respectively.  We use ${ReLU}(\mathbf{v})=\max (\mathbf{v}, \mathbf{0})$ to denote the ReLU activation function where $\max (\cdot)$ function is applied entry wise to a vector.

\subsubsection{Transformers}
For descriptive purposes, we use single head attention to illustrate. \zdiffminor{Moreover, we use $\mathbf{X} \in \mathbb{R}^{d \times n}$ to describe the input shape in the subsequent explanations, instead of $\mathbf{X} \in \mathbb{R}^{n \times d}$ depicted in Fig \ref{fig:figure1}}. \zdiffminor{We} use $\mathbf{Q}, \mathbf{K}$, and $\mathbf{V}$ to denote the multiplication of $\mathbf{W}_q\in \mathbb{R}^{d \times d}, \mathbf{W}_k\in \mathbb{R}^{d \times d}, \mathbf{W}_v\in \mathbb{R}^{d \times d}$, and $\mathbf{X} \in \mathbb{R}^{d \times n}$, \wq{ where $d$ represents the embedding dimensions, and $n$ represents the number of patches}. Therefore, we have:
\begin{equation}
\mathbf{Q}=\mathbf{W}_q \mathbf{X}, \quad \mathbf{K}=\mathbf{W}_k \mathbf{X}, \quad \mathbf{V}=\mathbf{W}_v \mathbf{X}
\label{eq:QKV}
\end{equation}

\zdiff{Then, a self-attention operation of input token sequence $\mathbf{X}\in \mathbb{R}^{d \times n}$ can be represented as
\begin{equation}
\operatorname{Att}(\mathbf{X})=\mathbf{V} \sigma\left(\mathbf{K}^{\top} \mathbf{Q}\right)
\label{eq:Att}
\end{equation}
}
The normalizing factor of $\frac{1}{\sqrt{d_{k q}}}$ is subsumed in the weight matrices $\mathbf{W}_k$ for notational simplicity. And we omitted the expression of multihead.

With \eqref{eq:QKV} and \eqref{eq:Att}, a standard Transformer layer $\tau$ can be represented as
\begin{equation}
\begin{aligned}
\operatorname{MLP}(\mathbf{X}) & =\left[\mathbf{W}_2 \operatorname{ReLU}\left(\mathbf{W}_1 \mathbf{X}_{:, 1}+\mathbf{b}_1\right)+\mathbf{b}_2+\mathbf{X}_{:, 1},\right. \\
& \left.\ldots, \mathbf{W}_2 \operatorname{ReLU}\left(\mathbf{W}_1 \mathbf{X}_{:, n}+\mathbf{b}_1\right)+\mathbf{b}_2+\mathbf{X}_{: n}\right] \\
\tau(\mathbf{X}) & =\operatorname{MLP}(\operatorname{Att}(\mathbf{X})+\mathbf{X}) .
\end{aligned}
\end{equation}

\subsubsection{Visual Prompt Tuning}
Visual prompts are learnable parameters prepended to other tokens before a Transformer encoder layer. These visual prompts $\mathbf{P} \in \mathbb{R}^{d \times n_p}$ have the same dimension as other tokens, where $n_p$ is the prompt length. Then the Transformer layers are represented as
\begin{equation}
\begin{aligned}
{\left[\mathbf{Z}^1, \mathbf{X}^1\right] } & =\tau^1\left(\left[\mathbf{P}, \mathbf{X}^0\right]\right) \\
{\left[\mathbf{Z}^i, \mathbf{X}^i\right] } & =\tau^i\left(\left[\mathbf{Z}^{i-1}, \mathbf{X}^{i-1}\right]\right) \quad i=2,3, \ldots, N \\
\mathbf{y} & =\operatorname{Head}\left(\mathbf{{X}}^N_{:,0}\right),
\end{aligned}
\end{equation}
where $\mathbf{Z}^i \in \mathbb{R}^{d \times n_p}$ represents the prompts computed by the $i$-th Transformer layer $\tau^i$, and $\mathbf{{X}}^N_{:,0}$ represents the CLS token.

\subsection{Embedded Prompt Tuning}
\xsh{VPT prepends prompt tokens before all original input tokens parallely, while VP adds the values of prompt tokens with original input tokens across all channels. These prompt introducing ways either do not fine-tune the original input tokens fine-grainedly or significantly disrupt the information of original input tokens. Therefore, we propose Embedded Prompt Tuning (EPT), embedding prompt tokens into expanded channels. In this way, EPT can not only preserve original information well, but also introduce extra useful context to optimize embedding tokens.}

As shown in \wq{Fig} \ref{fig:figure1}, the concept of our proposed EPT method and its differences compared to VPT are illustrated. We use the EPT prompts $\mathbf{P_E}$ to differentiate $\mathbf{P}$. $\mathbf{P_E} \in \mathbb{R}^{d_p \times n}$ have a completely different shape from the prompts $\mathbf{P} \in \mathbb{R}^{d \times n_p}$ in VPT, \wq{where $d_p$ represents the prompt length}. It should be noted that \zdiff{the prompts $\mathbf{P_E}$ are not directly inserted into the patch tokens but only appear during the softmax operation on $\mathbf{{K^T}Q}$ in the attention calculation, and they disappear after the computation is completed. The purpose of this approach is to scale $\mathbf{{K^T}Q}$ without altering its shape, enabling subsequent computations}.

Next, we will provide the formal formulation of EPT. Embedded prompts are learnable parameters embedded in the $\mathbf{{K^T}Q}$ matrix and jointly participate in subsequent softmax calculations. And these embedded prompts $\mathbf{P_E} \in \mathbb{R}^{d_p \times n}$ no longer participate in subsequent calculations after completing the softmax computation, where $d_p$ is the prompt length. Therefore, the attention mechanism is defined as

\begin{equation}
\operatorname{Att}\left(\mathbf{X};\mathbf{P_E}\right)=\mathbf{V} \sigma\left(\left[\begin{array}{c}
\mathbf{P_E} \\
\mathbf{K^{\top} Q}
\end{array}\right]\right)_{d_p:,:}
\label{eq:EPT1}
\end{equation}
where $\sigma\left(\left[\begin{array}{c}
\mathbf{P_E} \\
\mathbf{K^{\top} Q}
\end{array}\right]\right)_{d_p:,:}$ represents selecting values from the matrix $\mathbf{{K^T}Q}$ row-wise, starting from the $d_p$-th row up to the $(n+d_p-1)$-th row, after undergoing the prompt and softmax operations, thus maintaining its shape in subsequent calculations.

Then a single Transformer layer is represented as
\begin{equation}
\begin{aligned}
& \tau\left(\mathbf{X}; \mathbf{P_E}\right)=MLP\left(\operatorname{Att}\left(\mathbf{X}; \mathbf{P_E}\right)+\mathbf{X}\right)
\end{aligned}
\label{eq:EPT2}
\end{equation}

Two variations are available within the EPT method: EPT-SHALLOW and EPT-DEEP, distinguished by how extensively Transformer layers are utilized.

\paragraph{$\textbf{EPT-SHALLOW}$}
Only the first layer of the Transformer is embedded with prompts, therefore all the Transformer layers of EPT-SHALLOW are represented as

\begin{equation}
\begin{aligned}
\mathbf{X}^1 & =\tau^1\left(\mathbf{X}^0;\mathbf{P_E}^0\right) \\
\mathbf{X}^i & 
=\tau^i\left(\mathbf{X}^{i-1}\right) \quad i=2,3, \ldots, N \\
\mathbf{y} & =\operatorname{Head}\left(\mathbf{{X}^N}_{:,0}\right),
\end{aligned}
\end{equation}

\paragraph{$\textbf{EPT-DEEP}$}
\zdiff{All Transformer layers are embedded with prompts, and the Transformer layers of EPT-DEEP are represented as
\begin{equation}
\begin{aligned}
\mathbf{X}^1 & =\tau^1\left(\mathbf{X}^0;\mathbf{P_E}^0\right) \\
\mathbf{X}^i & 
=\tau^i\left(\mathbf{X}^{i-1};\mathbf{P_E}^{i-1}\right) \quad i=2,3, \ldots, N \\
\mathbf{y} & =\operatorname{Head}\left(\mathbf{{X}^N}_{:,0}\right),
\end{aligned}
\end{equation}
}

\paragraph{$\textbf{EPT Enables Patch-Wise Scaling}$}
For a matrix $\mathbf{K^T Q}$, we select the vector from the \wq{$i$}-th column to illustrate. Let $\mathbf{u}=\mathbf{{K^T}Q}_{:,i}=(\mathbf{{K^T}Q})_{:,i}$, which corresponds to the multiplication of the query of the \wq{$i$}-th patch and the keys of all patches. Then, the softmax value of $\mathbf{u}$ can be represented as
\begin{equation}
\sigma(\mathbf{u})=\frac{e^\mathbf{u}}{\left\|e^\mathbf{u}\right\|_1}=\quad\frac{1}{\sum_{i=1}^n e^{u_i}}\left[e^{u_1}, e^{u_2}, \cdots, e^{u_n}\right]
\label{eq:origin_u}
\end{equation}

For the prompted matrix $\left[\begin{array}{c}\mathbf{P_E} \\ \mathbf{K^{\top} Q}\end{array}\right]$, let $\mathbf{u_p}=\left[\begin{array}{l}\mathbf{p} \\ \mathbf{u}\end{array}\right] \in \mathbb{R}^{{d_p}+\wq{n}}$ be the \wq{$i$}-th column of the prompted matrix, where $\mathbf{p} \in \mathbb{R}^{d_p}$ and $\mathbf{u} \in \mathbb{R}^{\wq{n}}$ represent the prompt and the original vector respectively. After the softmax operation, the value of the $i$-th column vector of the original matrix becomes
\begin{equation}
\sigma(\mathbf{u_{p}})_{d_p:}=\frac{e^\mathbf{u}}{\left\|e^{\mathbf{u_{p}}}\right\|_1}=\quad\frac{1}{\sum_{i=1}^n e^{u_i}+\sum_{j=1}^{d_p} e^{p_j}}\left[e^{u_1}, e^{u_2}, \cdots, e^{u_n}\right]
\label{eq:scale_u}
\end{equation}

From equations \eqref{eq:origin_u} and \eqref{eq:scale_u}, we can observe that the softmax operation applied to the prompt enables the adjustment of each patch-level feature, corresponding to patch-wise scaling. It is worth noting that the dimensions of the input and output remain unchanged during the softmax operation. The advantages of patch-wise scaling will be further analyzed and compared in \wq{Sec \ref{sec:interpre}}.

\subsection{Analysis of EPT}
\subsubsection{Intuition}
In the context of few-shot learning, researchers face the challenge of overfitting due to an insufficient number of training samples, which is characterized by accurate classification on the training set but poor performance on the test set. To tackle \zdiffminor{this challenge}, we attempt to approach \zdiffminor{it from the following perspective: calibrating the distribution of the few-shot samples by reducing the intra-class distance of samples from the same class, thereby increasing the degree of separation between features of different classes during training, ultimately achieving higher accuracy \cite{zhang2023understanding}.}

\xsh{Based on our observations, EPT can help mitigate the negative impact caused by foundation models when learning pretraining data distribution. Additionally, \zdiff{we notice that the patch-wise scaling operation, by implementing varying scaling ratios for different samples, facilitates the aggregation of features from the same class towards the intra-class center. Consequently, it increases the potential for greater separation of features between samples from different classes.} Moreover, by introducing prompts, the dimensions of features are increased, thereby enhancing the high-dimensional expression and decoupling capability of features. Through analyzing these two operations intuitively and theoretically, we propose a new perspective to understand prompt tuning: Prompt is a distribution calibrator.}

\subsubsection{Interpretation}
\label{sec:interpre}
First, we define the \wq{intra-class} distance for a better understanding of the problem. For a dataset $\mathbf{X}=\left\{\mathbf{x}_{k, i}, k \in[K], i \in\left[n_k\right]\right\}$ with K classes, where $\mathbf{x}_{k, i} \in \mathbb{R}^d$ is the $i$-th sample in the $k$-th class, and the number of samples $n_k$ in each class is balanced with $n_1=\cdots=n_K=n$. \wq{We borrowed the definition from Yaras et al.~\cite{yaras2023law}, and we use $\bar{\mathbf{x}}_k$ to represent the center of the samples in class k, where the intra-class distance matrix is defined as}
\begin{equation}
\Sigma_\mathbf{W}=\frac{1}{N} \sum_{k=1}^K \sum_{i=1}^n\left(\mathbf{x}_{k, i}-\bar{\mathbf{x}}_k\right)\left(\mathbf{x}_{k, i}-\bar{\mathbf{x}}_k\right)^{\top}
\end{equation}

According to the definition, a smaller value of $\Sigma_\mathbf{W}$ indicates a higher degree of feature clustering, \wq{which may lead to better separability}.

\wq{
For dataset that satisfies $\mathbf{X}=\left\{\mathbf{x}_{k, i} \geqslant 0, k \in[K], i \in\left[n_k\right]\right\}$, If we apply different scaling operations to samples of the same class, we can derive the conclusion stated in Lemma \ref{lemma}.
\begin{lemma}
Let $c_{k, i}, c_{k, j} \in[0,1]$ be the scaling operations on the samples $\mathbf{x}_{k, i}$ and $\mathbf{x}_{k, j}$ of class k, respectively, and satisfy $\left(c_{k, i}-c_{k, j}\right)\left(\left\|\mathbf{x}_{k,i}\right\|_2-\left\|\mathbf{x}_{k,j}\right\|_2\right) \leqslant 0$, and  $\left(\left\|c_{k, i} \mathbf{x}_{k, i}\right\|_2-\left\|c_{k, j} \mathbf{x}_{k, j}\right\|_2\right)\left(\left\|\mathbf{x}_{k, i}\right\|_2-\left\|\mathbf{x}_{k, j}\right\|_2\right) \geqslant 0$. Then, the new intra-class distance matrix and the original intra-class distance matrix of class k satisfy $\Sigma_{\mathbf{W}}^{\prime} \leqslant {c_k}^2 \Sigma_\mathbf{W}$, where $c_k$ is the scaling factor of the $k$-th class center.
\label{lemma}
\end{lemma}
}
Proof of Lemma \ref{lemma} is presented in Appendix. Lemma \ref{lemma} demonstrates that a scaling operation $c_{k,i}$ that induces intra-class clustering more effectively relative to applying the same scaling factor to all samples within a class. The actual effect is that the farther the sample $\mathbf{x}_{k,i}$ is from the origin, the smaller the scaling factor it has, and vice versa. The advantage is that it can bring the samples closer to the intra-class center.

Furthermore, we can observe that \zdiff{by embedding prompts in the softmax function in EPT, we can achieve the scaling functionality defined in Lemma \ref{lemma}, thereby increasing the intra-class concentration}. Therefore, we conducted an analysis of the mechanism behind EPT. Considering the complex structure of the Transformer and the dataset, to simplify the problem, \textbf{we assume the number of patches (including the CLS token) is 2} and proceed with the analysis based on this assumption. 

\wq{Let $\mathbf{u}_1$ and $\mathbf{u}_2$ be the first column vectors of the $\mathbf{K^T Q}$ matrix for two different samples $\mathbf{X}_{k, 1}\in \mathbb{R}^{d\times n}$ and $\mathbf{X}_{k, 2}\in \mathbb{R}^{d\times n}$ within the same class k, which denote the CLS token, and \textbf{n=2}. Let $\mathbf{u}_{p,1}$ and $\mathbf{u}_{p,2}$ correspond to the vectors after the prompt. Then, we can draw the following conclusion.
\begin{proposition}
Consider an original matrix $\sigma\left(\mathbf{K}^{\top} \mathbf{Q}\right)$ and a prompted matrix $\sigma\left(\left[\begin{array}{c}\mathbf{P_E} \\ \mathbf{K^{\top} Q}\end{array}\right]\right)_{d_p:,:}$. We can always find a prompt $\mathbf{P}_{\mathbf{E}} \in \mathbb{R}^{d_p \times n}$ such that, for the CLS tokens of  $\mathbf{X}_{k, 1}\in \mathbb{R}^{d\times n}$ and $\mathbf{X}_{k, 2}\in \mathbb{R}^{d\times n}$, when $\left\|\sigma\left(\mathbf{u}_1\right)\right\|_2 \leqslant\left\|\sigma\left(\mathbf{u}_2\right)\right\|_2$, then $\frac{\left\|\sigma\left(\mathbf{u}_{p,1}\right)_{d_p:}\right\|_2}{\left\|\sigma\left(\mathbf{u}_1\right)\right\|_2} \geqslant \frac{\left\|\sigma\left(\mathbf{u}_{p,2}\right)_{d_p:}\right\|_2}{\left\|\sigma\left(\mathbf{u}_2\right)\right\|_2}$.
\label{prop}
\end{proposition}
}
Proof of Proposition \ref{prop} is presented in Appendix. According to Lemma \ref{lemma} and Proposition \ref{prop}, \zdiff{we can conclude that EPT can achieve more compact scaling operations by embedding prompts in the softmax, thereby reducing intra-class distances and making samples of the same class more concentrated.}

\section{Experiment}
\subsection{Experiment Setup}
This article aims to investigate the performance of mainstream PEFT methods in few-shot medical classification tasks, \deleted{preliminarily explore the potential of our proposed LAP method}and emphatically validate the effectiveness of our designed EPT method, particularly by comparing the characteristics of VPT and similar PEFT methods to explore their commonalities and capabilities\deleted{ to strike a balance between fitting and generalization}. First, we will introduce datasets, tasks, and experimental settings of our study.

\paragraph{Datasets}
We evaluate on the MedFMC2023 grand challenge public classification dataset \cite{wang2023medfmc} which covers three different modalities: 
(1)Thoracic Disease Screening (Chest). 2,140 for training, 2,708 for validation, and 3,869 for testing.
(2)Pathological Tumor Tissue (Colon). 5,654 for training, 4,355 for validation and 7,651 for testing.
(3)Endoscopy Images for Lesion (Endo). 1,810 for training, 2,055 for validation and 2,936 for testing.

\begin{table*}[!ht]
\captionsetup{singlelinecheck=off, justification=raggedright,labelsep=period}
\caption{$\textbf{Image classification accuracy for ViT-Base/16}$. We reported mean mAP and Acc for 1-shot, 5-shot and 10-shot on the Chest, Colon, and Endo datasets individually, as well as the average \zdiffminor{accuracy score} for a total of 9 tasks across the three datasets and three shot numbers. \deleted{"Tuned/Total" is the max percentage of tuned parameters required by 9 tasks.} \wq{The percentage inside the parentheses refers to the percentage improvement of all fine-tuning methods relative to the Linear method.} \deleted{'Ours(224)' refers to an image size of 224, while 'Ours(384)' refers to an image size of 384.}}
\centering
\resizebox{\linewidth}{!}{
\begin{tabular}{c|ccc|ccc|ccc||c}
\hline
\multirow{2}{*}{\begin{tabular}{c} 
ViT-Base/16 \\
$(85.8 \mathrm{M})$
\end{tabular}} & \multicolumn{3}{c|}{Chest} & \multicolumn{3}{c|}{Colon} & \multicolumn{3}{c||}{Endo} & \multirow{2}{*}{\begin{tabular}{c} 
average
\end{tabular}} \\

&1-shot & 5-shot & 10-shot &1-shot & 5-shot & 10-shot &1-shot & 5-shot & 10-shot\\
\hline \hline 
Full & 10.61 &11.16 &11.99&63.90 & 77.02 & 83.85 & 14.05 & 15.66& 17.57& 33.98(-9.34\%)\\
\hline 
\rowcolor{lightgray} Linear& 11.93 &14.88 &16.75&70.24 & 81.22 & 85.23 & 15.60 & 20.18& 21.30& 37.48(0.00\%)\\
Partial-1& 11.93 &14.48 &16.83&73.11 & 82.55 & 85.98 & 16.62 & 20.86& 24.05& 38.49(+2.69\%)\\
MLP-3 & 12.54 &15.85 &17.78&70.56 & 81.63 & 86.13 & 16.34 & 20.29& 21.67& 38.09(+1.63\%)\\
\hline 
LoRA& 12.30 &15.66 &17.40&71.91 & 81.69 & 84.88 & 16.01 & 20.12& 21.37& 37.93(+1.20\%)\\
Bias & 12.12 &15.00 &17.76&71.68 & $\mathbf{83.58}$ & 86.59 & 16.68 & 20.23& 21.66& 38.36(+2.35\%)\\
Adapter& 12.11 &15.32 &17.36&73.79 & 83.03 & 85.52 & 16.86 & 22.51& 24.95& 39.05(+4.19\%)\\
\hline 
VP& 12.22 &15.65 &17.33&73.70 & 81.26 & 85.11 & 15.67 & 19.46& 20.79& 37.91(+1.15\%)\\
VPT& 12.15 &14.88 &17.54&74.44 & 82.15 & 86.25 & 17.60 & 21.27& 22.89& 38.79(+3.50\%)\\
\rowcolor{white} Ours & $\mathbf{12.73}$ &$\mathbf{16.16}$ &$\mathbf{18.09}$&$\mathbf{75.31}$ & 83.34 & $\mathbf{87.23}$ & $\mathbf{17.66}$ & $\mathbf{22.80}$& $\mathbf{25.07}$& $\mathbf{39.82(+6.24\%)}$ \\
\hline
\end{tabular}\label{tab:vb16}
}
\end{table*}

\paragraph{Tasks} Each task is formulated as N-Way K-shot classification tasks. Due to the backbone network is pre-trained on ImageNet-21k, which owns a domain gap with medical images, the classification tasks are cross-domain few-shot classification tasks. The few-shot medical classification task, especially the multi-label classification task, particularly highlights the generalization performance of these PEFT methods. We mainly focus on few-shot scenerios rather than fully supervised training, and K is set to 1,5,10. 1-shot refers to every N images that sample from each class, rather than one image. Among the three datasets, ChestDR and Endo are multi-label classification tasks with 19 and 4 categories respectively, while Colon is a single-label classification task with 2 categories.

Accuracy and mean average precision are used to evaluate both single-label classification tasks and multi-label classification tasks. We compare widely used PEFT methods with our baseline method VPT. To reduce the error caused by random sampling, we follow the setup of MedFMC and performed 5 random samplings. We calculate the average accuracy score for four runs for each sampling.
\begin{figure*}[!htbp]
  \centering
  \includegraphics[width=1\textwidth]{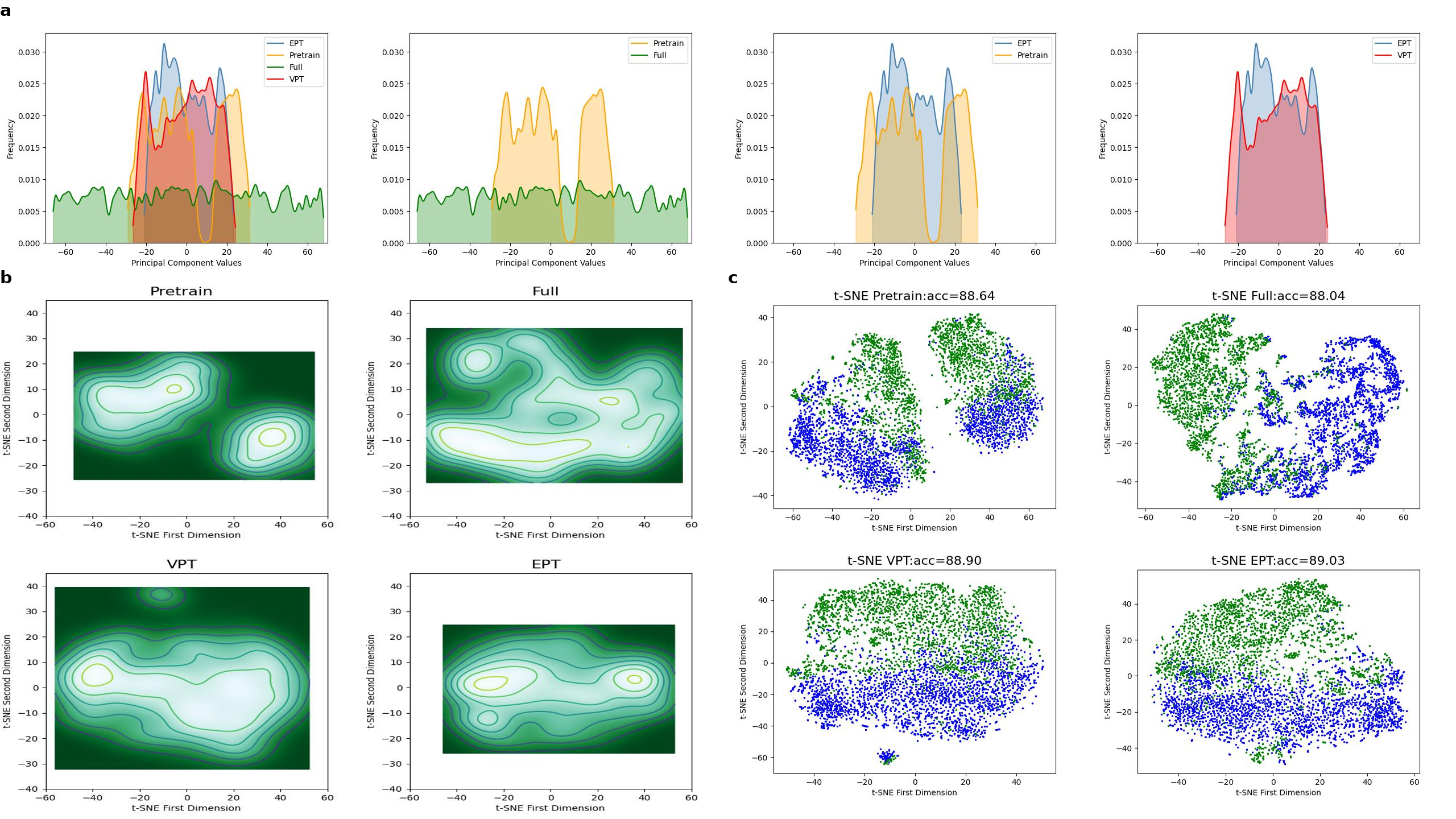}
  \captionsetup{singlelinecheck=off, justification=raggedright,labelsep=period} \caption{$\textbf{Visual Analysis of Feature Distribution of Colon Test Dataset}$. \textbf{a.} \wq{Principal component distribution results of features from the same class in the Colon test dataset after training with Full, Linear, VPT, and EPT methods.} \textbf{b.} \wq{Contour visualization of two-dimensional projections of features from the same class in the Colon test dataset after training with Full, Linear, VPT, and EPT methods.} \textbf{c.} \wq{T-SNE visualization of two-dimensional projections of features from different classes in the Colon test dataset after training with Full, Linear, VPT, and EPT methods.}}
  \label{fig:feature_sep}
\end{figure*}

\paragraph{Baselines}
\zdiff{We follow MedFMC and compare EPT with other widely applied PEFT methods}. We compare the results of \deleted{two}Vision Transformer architecture\deleted{, Vision Transformer} \cite{dosovitskiy2020image} (ViT)\deleted{ and Swin Transformer(Swin)}, on image classification in Sec \ref{sec:6.2}. Due to the fact that the EPT method is also a prompt-based fine-tuning approach, it shares many similarities with VPT and VP. Additionally, given the superior performances of VPT and VP, we primarily compare our results with VPT and VP. Unlike selecting the Full method as the primary reference in \cite{jia2022visual}, our experiments chose the Partial-1\cite{yosinski2014transferable} fine-tuning method as the primary reference. The reason is that for few-shot fine-tuning, fine-tuning all parameters incurs computational costs that do not match the data volume, and our experiments have found that the Full method often yields the worst results. Therefore, we chose Partial-1 as the primary reference method. \deleted{For Adapter\cite{chen2022adaptformer} and Lora\cite{hu2021lora}, we set the dimension $r$ for downsampling to 4.} \zdiff{The default prompted layers for EPT and VPT are both set to 12, meaning the prompt is used through all the layers. The prompt length for prompt-based fine-tuning methods is set to achieve the best performance individually. If not otherwise specified, we adopt ViT-Base/16 pretrained on ImageNet-21k as our default backbone model, which is also the default model for ViT in the MedFMC Challenge}.

\deleted{For the two relative embedding orders of prompts in the EPT method: EPT and EPT-C, we select different relative embedding orders for each task and choose the best result as the final result of our approach. We do this because we found that EPT-C often exhibits better generalization performance, while EPT tends to have better fitting ability. This leads to each method being more suitable for relatively simple datasets (Colon) and relatively complex datasets (Chest) respectively, and this difference brings about significant performance superiority for our method. Furthermore, we conducted a more in-depth analysis of this phenomenon in Sec \ref{sec:6.3}. In terms of fairness, our standalone EPT-C or EPT can surpass most other PEFT methods, especially the VPT that we specifically compared against. More results and details are provided in Sec \ref{sec:6.2} and Appendix.}

\begin{table*}[!ht]
\captionsetup{singlelinecheck=off, justification=raggedright,labelsep=period}
\caption{\zdiffminor{$\textbf{Different ViT backbone scales.}$ $\textbf{Image classification accuracy for ViT-Large/16}$ pretrained on ImageNet-21k. We reported the average mAP and Acc for 1-shot, 5-shot and 10-shot on the Chest, Colon, and Endo datasets individually, as well as the average accuracy score for a total of 9 tasks across the three datasets and three shot numbers. Best results among all methods are \textbf{bolded}. The percentage inside the parentheses refers to the percentage improvement of all fine-tuning methods relative to the Linear method.}}
\centering
\resizebox{\linewidth}{!}{
\begin{tabular}{c|ccc|ccc|ccc||c}
\hline
\multirow{2}{*}{\begin{tabular}{c} 
ViT-Large/16 \\
$(303.7 \mathrm{M})$
\end{tabular}} & \multicolumn{3}{c|}{Chest} & \multicolumn{3}{c|}{Colon} & \multicolumn{3}{c||}{Endo} & \multirow{2}{*}{\begin{tabular}{c} 
average
\end{tabular}} \\

&1-shot & 5-shot & 10-shot &1-shot & 5-shot & 10-shot &1-shot & 5-shot & 10-shot\\
\hline \hline 
Full & 10.75 &11.17 &11.84&70.80 & 82.19 & 86.22 &17.22 &20.06 &21.72& 36.88(-5.94\%)\\
\hline 
\rowcolor{lightgray} Linear&13.19	&16.67	&18.44	&73.65	&80.04	&84.49	&19.08	&23.47	&23.87	&39.21(0.00\%)\\
Partial-1&13.27	&17.05	&19.12	&71.65	&81.53	&85.98	&18.54	&$\mathbf{24.70}$	&26.54	&39.82(+1.56\%)\\
MLP-3 &11.49 &12.24 &13.47 &66.12 &83.83 &$\mathbf{89.53}$ &15.15 &17.25 &20.64 &36.64(-6.55\%)\\
\hline 
LoRA &13.38	&16.80	&18.41	&75.00	&81.04	&84.82	&$\mathbf{19.62}$	&23.94	&23.40 &39.60(+0.99\%)\\
Bias &13.30	&17.28	&18.60	&73.86	&81.20	&86.28	&18.71	&22.47	&23.05	&39.42(+0.54\%)\\
Adapter&11.12	& 12.53	& 15.06	& $\mathbf{76.53}$	& 83.38	& 86.67	& 16.76	& 22.51	& 24.73	& 38.81(-1.02\%)\\
\hline 
VP &13.30	&17.30	&18.68	&73.31	&85.54	&87.36	&18.48	&23.00	&23.39	&40.04(+2.12\%)\\
VPT &13.20	&16.05	&18.41	&73.74	&80.90	&86.06	&18.73	&23.91	&$\mathbf{26.74}$ &39.75(+1.38\%)\\
\rowcolor{white} Ours &$\mathbf{13.73}$	&$\mathbf{17.67}$	&$\mathbf{19.53}$	&76.33	&$\mathbf{85.91}$	&87.53	&18.93	&23.38 &23.78	&$\mathbf{40.75(+3.93\%)}$ \\
\hline
\end{tabular}\label{tab:vl16}
}
\end{table*}

\subsection{Main Results}\label{sec:6.2}

\paragraph{EPT on MedFMC} We present average results of various PEFT methods for 1-shot, 5-shot and 10-shot classification on three datasets: Chest, Colon and Endo, as shown in Tab \ref{tab:vb16}. Several main findings can be observed. 

\xsh{First, EPT outperforms other PEFT methods by a substantial margin on three datasets in most cases. Specifically, \zdiff{it surpasses the second-best method Adapter by 2.05\% and the third-best method VPT by 2.74\%, respectively}. EPT still achieves second place in the Colon 5-shot task, with a performance gap of only 0.24 from Bias. This demonstrates that the superior performance of EPT when adapting pre-trained foundation models to cross-domain few-shot scenarios, e.g., medicial image analysis. Second, in the category of prompt tuning, EPT surpasses VPT and VP by 2.74\% and 5.09\%, respectively. \zdiff{In addition, EPT outperforms LoRA by 5.04\%}. This not only indicates that EPT has addressed previous shortcomings in the way of introducing prompts, embedding prompts in the channel direction is a more promising approach. Moreover, EPT has better approximation capabilities on Transformer architectures, thereby breaking through limitations of prompt tuning. \zdiff{Third, EPT, VPT and VP surpasses Linear by 6.24\%, 3.50\%, and 1.15\%, respectively}, and prompt tuning achieves very competitive ranks among all PEFT methods. This shows that prompt tuning helps to alleviate the negative impact of foundation models in learning pre-training data distributions, and calibrates their performance in new domains. Fourth, Full is 9.34\% lower than Linear, potentially due to overfitting. When adapting foundation models to few-shot scenarios, fine-tuning more parameters is not always better, and PEFT methods are very meaningful.}

\paragraph{Analysis of Feature Distribution}
\xsh{As shown in Fig \ref{fig:feature_sep}, we present visualization results of feature distribution after fine-tuning on the Colon dataset.}

\xsh{In Fig \ref{fig:feature_sep} (a), it displays the feature distribution of samples from the same class on the first principal component obtained by principal component analysis (PCA). We select 2,628 samples labeled as existing tumours, and four methods were compared: Full, Linear(Pretrained backbones), VPT, and EPT. The x-axis represents feature values, and the y-axis represents the number of samples. Results show that: (1) EPT exhibits a narrower range of sample distribution, with more pronounced feature values, demonstrating a more concentrated intra-class feature distribution. (2) Foundation models only pre-trained on natural images displays a clear different behavior, and lack continuity, suggesting that samples may cluster into two distinct groups that are significantly far from each other. (3) Full displays that samples are overly dispersed and exhibit lower feature values, which could be due to overfitting and the failure of capturing common features.}

\begin{figure*}[h]
  \centering
  \includegraphics[width=1\textwidth]{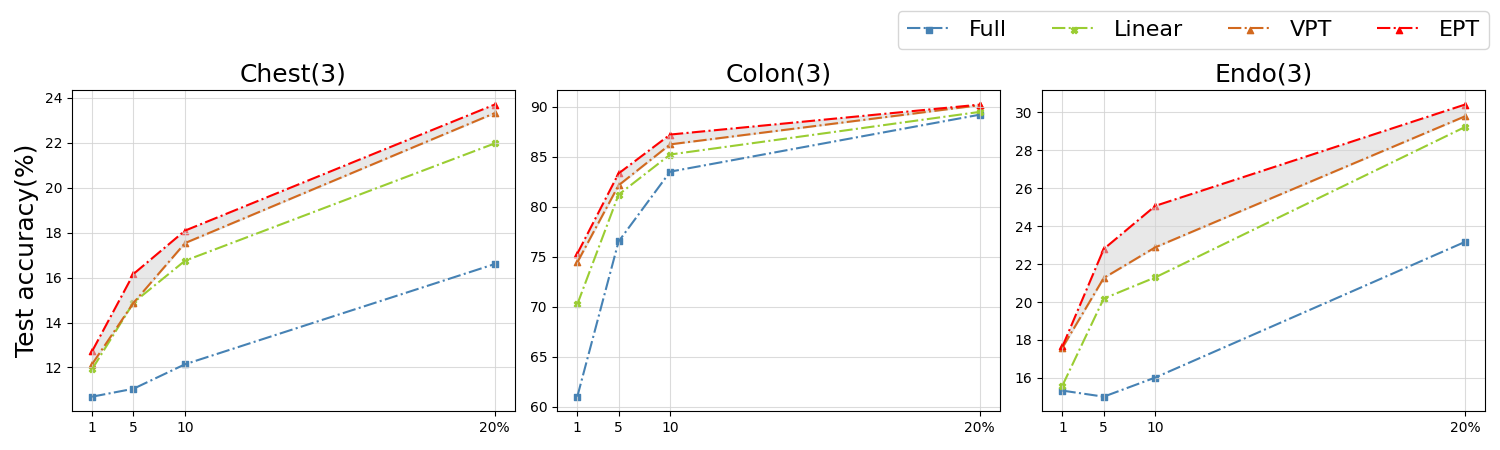}
  \captionsetup{singlelinecheck=off, justification=raggedright,labelsep=period} \caption{$\textbf{Comparative Analysis Across Diverse Downstream Data Volumes}$. We report the performance of three methods on 1-shot, 5-shot, 10-shot, and 20\% of the training set. The highlighted regions demonstrate the accuracy differences between EPT and VPT. All methods use ViT-Base/16 as the backbone\deleted{, and the image size for EPT is 384}.}
  \label{fig:data_scale}
\end{figure*}

\xsh{In Fig \ref{fig:feature_sep} (b), it visualizes the two-dimensional feature distribution of samples from the same class by t-SNE. Our selcted samples and methods are same as Fig \ref{fig:feature_sep} (a). The x-axis and y-axis represent feature values of two different dimensions, respectively. Results show that: (1) Pretrain shows that the density contour lines on the left and right sides do not connect and close, indicating the discontinuity and a huge distance within the distribution. (2) Full shows that the density contour lines are irregular, indicating that samples are dispersed throughout the space, and not sufficiently concentrated. (3) VPT shows a clear outlier distribution on the top, far from the center. (4) EPT shows that the area enclosed by the density contour lines is the smallest, indicating that samples are more concentrated. \zdiff{Fig \ref{fig:feature_sep} (a) and Fig \ref{fig:feature_sep} (b) both illustrate that EPT has a more continuous and concentrated intra-class feature distribution. This may be due to the patch-wise scaling operation, which reduces the distribution scopes and extracts more universal and effective features.}}

\xsh{In Fig \ref{fig:feature_sep} (c), it displays the two-dimensional feature distribution of all classes by t-SNE. We randomly select 5,256 samples from two categories, maintaining a 1:1 ratio. Results show that: (1) Pretrain not only exhibits a larger intra-class distribution but also shows inter-class linear inseparability. (2) Prompt tuning demonstrates a continuouts and concentrated intra-class distribution, and the separability between classes are also distinct. \zdiff{This may be due to the introduction of prompt tokens, which expand the dimensionality of the input space, thereby increasing the separability of features.}}

\xsh{\zdiff{These visualization results provide strong evidence for the view that prompt tuning acts as a distribution calibrator.}}

\paragraph{EPT on Larger Backbone Scale}
\zdiffminor{For the purpose of observing the performances of various tuning methods at larger backbone scales, we conducted experiments on ViT-Large/16. The experimental results, as shown in Tab \ref{tab:vl16}, indicate that our EPT method still demonstrates the most significant advantages on all three datasets and outperforms other methods. Additionally, we can observe that fine-tuning with ViT-Large/16 as the backbone generally yields better classification performance compared to ViT-Base/16. Even linear methods show improved classification results. However, both MLP-3 and Adapter methods exhibit a decrease in performance, which differs from the results in Tab \ref{tab:vb16}. Furthermore, the average classification results of EPT demonstrate an increasing trend compared to Tab \ref{tab:vb16}. This proves the effectiveness of our EPT method on larger-scale pre-trained models.}

\paragraph{EPT on Different Data Scale} 
\xsh{We vary the training data from 1, 5, 10-shot to 20\% of the entire training set. The average accuracy of different fine-tuning methods on various data scales is presented in Fig \ref{fig:data_scale}. First, we observe that our method consistently outperforms others across all data scales, displaying superior data scalability. Moreover, as the data scale increases, all fine-tuning methods exhibit a significant increase in accuracy, but their performance growth rate tend to slow down. Further observation reveals that the performance on Colon converges faster than on Chest and Endo. This might be because that multi-label tasks require more data to learn sufficient discriminative information.}

\subsection{Abalation Study}\label{sec:6.3}
In this section, EPT is consistently evaluated using the default experimental settings: ViT-Base/16 as the backbone\added{.}\deleted{, and an image size of 384.} Except for the ablation study on prompt length, \replaced{\wq{the prompt length for prompt-based fine-tuning methods is set to achieve the best performance individually}.}{the default prompt length was set to 1.}

\begin{figure*}[!htbp]
  \centering
  \includegraphics[width=1\textwidth]{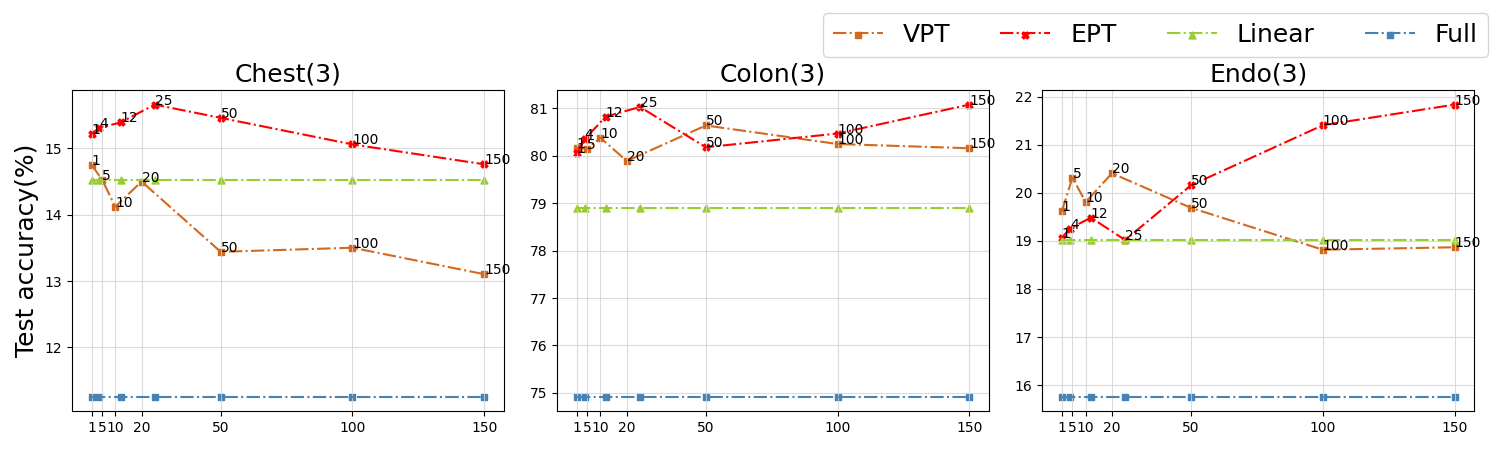}
  \captionsetup{singlelinecheck=off, justification=raggedright,labelsep=period} \caption{$\textbf{Ablation study on length of prompt}$. We vary the prompt length \wq{from 1 to 150}\deleted{1, 5, 10, 20, and 50} for EPT and VPT}
  \label{fig:length}
\end{figure*}

\begin{figure*}[!t]
  \centering
  \includegraphics[width=1\textwidth]{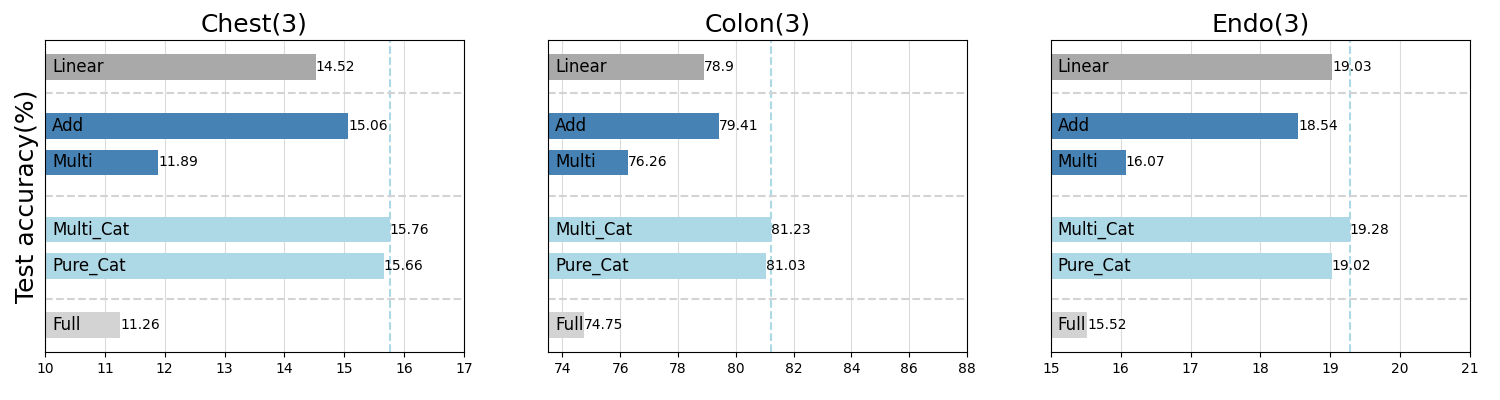}
  \captionsetup{singlelinecheck=off, justification=raggedright,labelsep=period} \caption{$\textbf{Ablation study on prompt embedding \wq{way}\deleted{Order}}$. We conducted experiments on MedFMC with four different \deleted{relative} embedding \wq{ways}\deleted{orders} of prompts.}
  \label{fig:location}
\end{figure*}

\begin{figure*}[!t]
  \centering
  \includegraphics[width=1\textwidth]{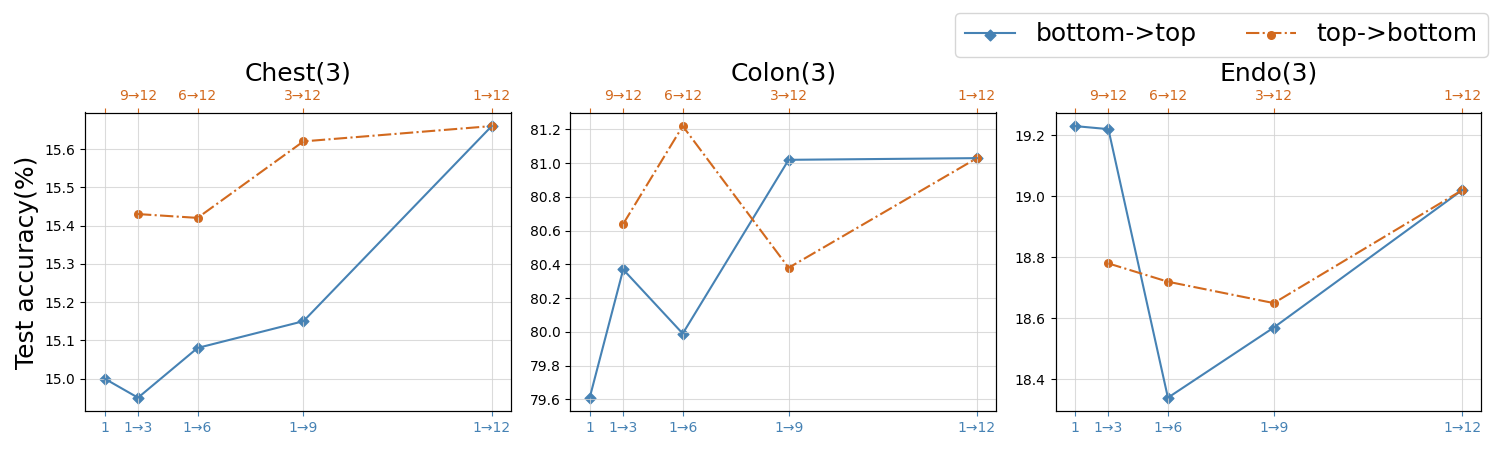}
  \captionsetup{singlelinecheck=off, justification=raggedright,labelsep=period} \caption{$\textbf{Ablation study on depth of prompt}$. We set the prompt length \wq{to be the same as in the main result}\deleted{to 1, which yielded the best results}, for each variant of EPT with validation sets. The indices i → j specify where prompts are embedded within the Transformer layers.}
  \label{fig:depth}
\end{figure*}

\paragraph{Length of Prompt}
\xsh{The average accuracy of different prompt tuning methods on various prompt length is presented in Fig \ref{fig:length}. We vary prompt length from 1, 5, 10, 20, 50, 100 to 150. Since EPT has a different way of introducing prompts from VPT, the number of prompt parameters from VPT is four times that of EPT when introducing a prompt with length of 1 in a single Transformer layer. Therefore, to maintain relatively consistent number of parameters, we use relative prompt length on the x-axis. This means when the relative prompt length is 1, the actual prompt length of VPT is 1, and the actual prompt length of EPT is 4.}

\xsh{As the prompt length increases, the performance of VPT generally declines across three datasets. In contrast, EPT shows a clear upward trend on Colon and Endo, and remains relatively stable on Chest. This suggests that EPT can better address overfitting issues. Moreover, when the prompt length is only 1, the overall performance of EPT and VPT is nearly identical. As the prompt length and the number of prompt parameters increase, yet remain within the scope of PEFT, EPT gradully begins to outperform VPT by a large margin. This indicates that EPT has better parameter scalability, with a much higher upper limit on approximation capabilities compared to VPT. We can also observe that both EPT and VPT can achieve better performance at shorter prompt length compared to longer ones in some cases. This suggests that longer prompt length is not always necessarily better, we can explore locally optimal solutions to balance performance and efficiency.}

\paragraph{Embedding Way of Prompt}
\xsh{The average accuracy of different EPT variants on various prompt embedding ways is presented in Fig \ref{fig:location}. \zdiff{We present four prompt embedding ways: add, multiply, concat purely, multiply and concat. "add" means the sum of ${P_E}$ and $K^\top Q$. "multiply" means the product of ${P_E}$ and $K^\top Q$. "concat purely" means the concatenation of ${P_E}$ and $K^\top Q$. "multiply and concat" is a novel EPT variant based on the insight of patch-wise scaling. We first obtain a scaling vector $\alpha \in \mathbb{R}^{n}$ by calculating difference between the maximum and minimum values in each patch of $K^\top Q$. Then, we multiply ${P_E}$ with $\alpha$, and concat the product with $K^\top Q$.}}

\xsh{\zdiff{The outcome shows that $"Multi\_Cat"$ achieves the best performance across three datasets. After applying the scaling vector $\alpha$, the values of prompts become larger, and original values become smaller according to eq \ref{eq:scale_u}}. Large original values are reduced more significantly, while small original values are reduced less. Therefore, the absolute distance between values decrease, enhancing characteristics of patch-wise scaling and further reducing the distance of intra-class feature distribution.\zdiff{ Moreover, $"Pure\_Cat"$ has better performance compared to $"Add"$ and $"Multi"$}, validating our analysis of prompt introducing ways again. \zdiff{The way of concatenation not only can helps to preserve the information of original tokens, but also introduces additional useful context in a more fine-grained manner.}}

\paragraph{Depth of Prompt}
\xsh{The average accuracy of EPT on various prompt depth and embedding orders is presented in Fig \ref{fig:depth}. Prompt depth is varying from 1 to 12, and embedding orders are varying from top to bottom (orange line) and bottom to top (blue line). Two valuable questions are mentioned below:}

\xsh{\textbf{Shallow or deep?} Overall, as the number of prompted layers increase, the performance of EPT improves. As the depth increases, prompts can acquire a more complex and abstract representation capability and adjust features from both shallow and deep layers.}

\xsh{\textbf{Top or bottom?} We can observe that the performance of EPT improves when embedding prompts from top to bottom, which is contrary to the conclusion of VPT. This may be related to prompt introducing ways. The foreground and background across different medical images are similar, which means there are fewer low-level and coarse-grained semantic features that are discriminative. Instead, high-level and fine-grained semantic features are more crucial. Since EPT directly embeds prompts in the channel direction, it can correlate with tokens more fine-grainedly and increase the dimension of representation. This results in greater enhancement of the high-level semantic features from deeper layers.}

\begin{figure*}[h]
  \centering
  \includegraphics[width=1\textwidth]{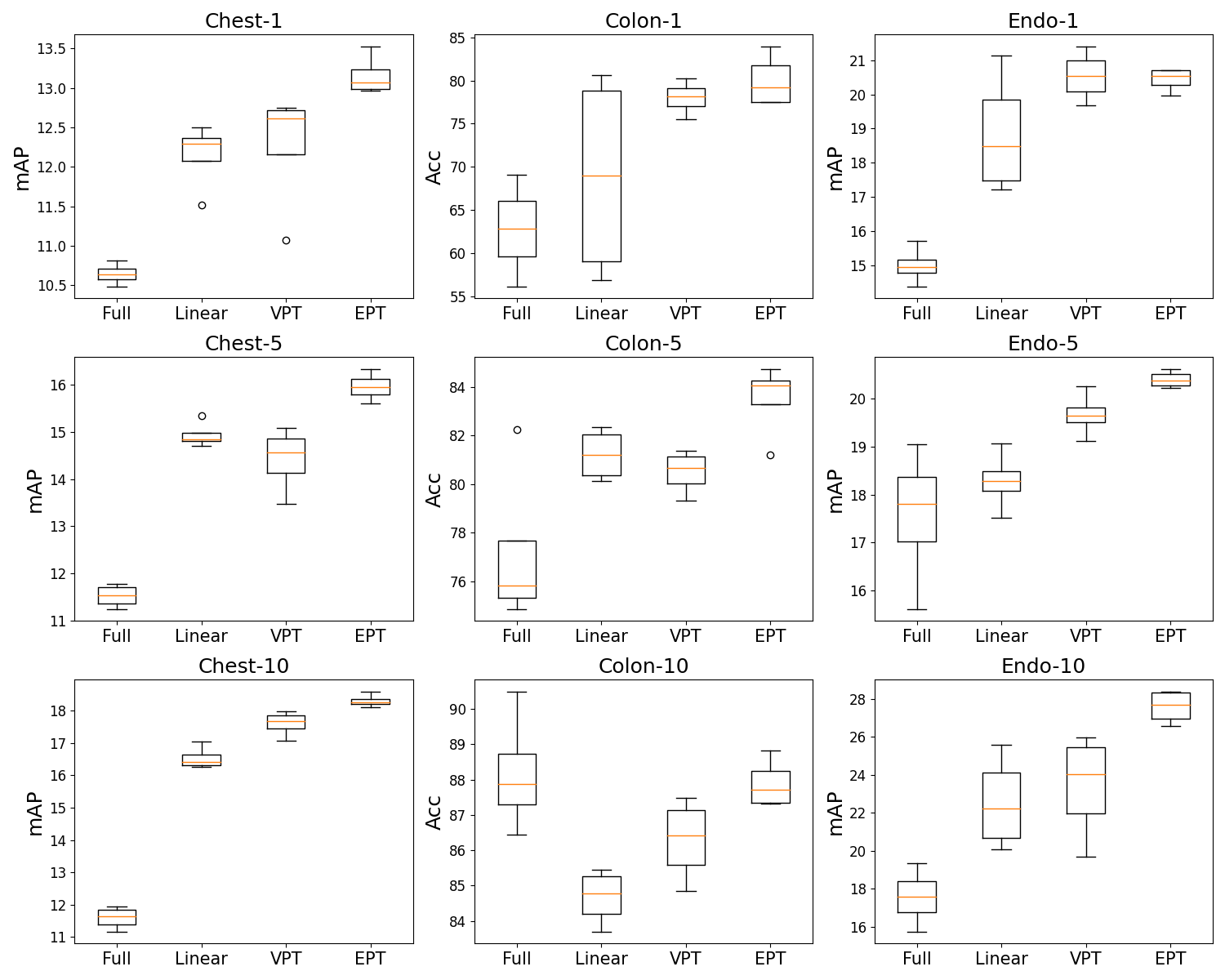}
  \captionsetup{singlelinecheck=off, justification=raggedright,labelsep=period} \caption{$\textbf{Visualization of convergence and distribution}$. For each task, we conduct 4 runs and compare the convergence and distribution of EPT with Full, Linear, and VPT.}
  \label{fig:box_plot}
\end{figure*}

\subsection{Futher Analysis}\label{sec:6.4}

In this section, we conduct a more in-depth analysis of \wq{VPT and} EPT. \wq{First, we performed an experiment by setting the prompt grad in VPT to false, which means completing the training with the prompt being frozen and only the linear head being trainable. Additionally, we also conducted a convergence experiment. As for the convergence experiment,} \deleted{We perform fitting experiments and convergence experiments. For both experiments,}we select a specific sampled dataset for training (provided in \cite{wang2023medfmc}). \deleted{For the fitting experiments, we use the same dataset for training and testing to evaluate the accuracy of various methods on the training set.}For the convergence experiment, we test methods on the default testing set and record the distribution of results, which reflects the convergence performance of the different methods

\paragraph{Set Prompt Grad False}
\xsh{The impact of grads of prompt parameters is shown in Tab \ref{tab:grad_false}. The outcome demonstrates that the performance of foundation models still improves in downstream tasks compared to models that are only pre-trained on natural images, even though we freeze the prompt parameters of VPT. This suggests that prompt tuning enhances the feature separability and strengthen the distribution calibration characteristics by introducing additional prompt tokens and increasing the input dimensions.}

\begin{table}[]
    \centering
    \captionsetup{singlelinecheck=off, justification=raggedright,labelsep=period} 
    \caption{\textbf{The impact of prompt grad on experimental results}. Let VPT-F and VPT-T represent prompt grad=false and prompt grad=true, respectively. Linear and VPT-T represent the percentage improvement in performance compared to VPT-F. The prompts are initialized \wq{the same}\deleted{with the same normal distribution}.}
\begin{tabular}{c||c|c|c}
\hline VPT-F & Chest & Colon & Endo \\
\hline \hline 1-shot & $12.25 $ & $72.48 $ & $17.29 $ \\
5-shot & $15.77 $ & $81.78 $ & $20.95 $ \\
10-shot & $17.62 $ & $85.62 $ & $22.52 $ \\
\hline
\hline Linear & Chest & Colon & Endo \\
\hline \hline 1-shot & $-2.63\% $ & $-3.09\%$ & $-9.78\% $ \\
5-shot & $-5.61\% $ & $-0.69\% $ & $-3.64\% $ \\
10-shot & $-4.91\% $ & $-0.45\% $ & $-5.41\% $ \\
\hline
\hline VPT-T & Chest & Colon & Endo \\
\hline \hline 1-shot & $-0.82\% $ & $+2.71\% $ & $+1.79\% $ \\
5-shot & $-5.69\% $ & $+0.45\% $ & $+1.53\% $ \\
10-shot & $-0.45\% $ & $+0.74\% $ & $+1.63\% $ \\
\hline
\end{tabular}
    \label{tab:grad_false}
\end{table}

\paragraph{Distribution of Convergence}
\xsh{The visualization results of convergence distribution are showned in Fig \ref{fig:box_plot}. \zdiff{Results indicate that EPT has the highest average values and the smallest variance in the convergence interval across various tasks, making it a superior and reliable PEFT method}. Moreover, both VPT and EPT achive better convergence than Full, proving the superiority of PEFT in cross-domain few-shot scenarios.}

\begin{table}[!ht]
\captionsetup{singlelinecheck=off, justification=raggedright,labelsep=period} \caption{\zdiffminor{$\textbf{Classification on ISIC 2018}$. {Image classification accuracy for ViT-Base/16} pretrained on ImageNet-21k. We reported the average Acc for 1-shot, 5-shot and 10-shot on the ISIC 2018 dataset. The percentage inside the parentheses refers to the percentage improvement of all fine-tuning methods relative to the Linear method.}}
\centering
\footnotesize
\begin{tabular}{p{2.2cm}||p{1.0cm}p{1.0cm}p{1.0cm}|p{2.5cm}}
\hline 

\multirow{2}{*}{\begin{tabular}{c} 
ViT-Base/16 \\
$(85.8 \mathrm{M})$
\end{tabular}}
 &  \multicolumn{3}{c|}{ ISIC 2018 } & \multirow{2}{*}{
average

}\\

& 1-shot & 5-shot & 10-shot & \\
\hline \hline 
Full &24.13	&31.95	&33.55	&29.87(-14.51\%)\\
\hline 
\rowcolor{lightgray}Linear & 25.74	&38.39	&40.68	&34.94(0.00\%)\\
 
Partial-1 & 28.69	&36.39	&40.80	&35.29(+1.00\%)\\
 
MLP-3 & 22.09	&37.91	&38.00	&32.67(-6.49\%)\\
\hline 
LoRA & 28.52	&38.39	&41.48	&36.13(+3.41\%)\\
Bias & 27.81	&37.75	&42.78	&36.11(+3.35\%)\\
Adapter & 25.09	 &38.47	 &44.59	 &36.05(+3.18\%)\\
\hline 
VP & 26.07	&36.54	&39.33	&33.98(-2.75\%)\\
VPT & 26.88	&39.48	&44.46	&36.94(+5.72\%)\\
\rowcolor{white} Ours & $\mathbf{28.89}$	&$\mathbf{41.35}$	&$\mathbf{44.65}$	&$\mathbf{38.30(+9.62\%)}$\\
\hline
\end{tabular}\label{tab:isic}
\end{table}

\zdiffminor{
\paragraph{EPT on More Classification Tasks}
For further exploration of the performance of fine-tuning methods in few-shot medical image analysis, we also conducted experiments on the ISIC 2018 dataset~\cite{codella2019skin}(see Appendix for details). From Tab \ref{tab:isic}, it can be observed that some fine-tuning methods failed, such as MLP-3 and VP, both exhibiting worse performance compared to the Linear method. However, methods targeting backbone fine-tuning, such as LoRA, Bias, and Adapter, still showed more significant improvements compared to Linear. VPT and EPT demonstrated even more notable enhancements, with EPT outperforming other fine-tuning methods by nearly 10\% points relative to Linear, thus confirming the effectiveness of the EPT method.}
\section{Conclusion}
\xsh{In this paper, we propose an effective PEFT method in cross-domain few-shot scenarios, e.g., medical image analysis, namely Embedded Prompt Tuning (EPT). EPT embeds prompt tokens into the expanded channels. In this way, EPT can introduce more useful context, while preserving the original information. Through analyzing patch-wise scaling and feature separation operations from EPT, we also find that prompt is a distribution calibrator, mitigating the negative impact caused by foundation models when learning feature distributions on pre-training data. Experiments show that our EPT can not only achieve superior performance, but also complete the fine-tuning process within competitive time. Considering limitations, there is still significant improvement space for EPT in fine-grained perception tasks, e.g., segmentation. In the future, we will continue to unleash the power of EPT on fine-grained perception tasks, and conduct a further theoretical exploration of the essence of EPT in distribution calibration.}

\bibliography{main}

\begin{thebibliography}{10}

\bibitem{bommasani2021opportunities}
Rishi Bommasani, Drew~A Hudson, Ehsan Adeli, Russ Altman, Simran Arora, Sydney von Arx, Michael~S Bernstein, Jeannette Bohg, Antoine Bosselut, Emma Brunskill, et~al.
\newblock On the opportunities and risks of foundation models.
\newblock {\em arXiv preprint arXiv:2108.07258}, 2021.

\bibitem{zhang2023text}
Yunkun Zhang, Jin Gao, Mu~Zhou, Xiaosong Wang, Yu~Qiao, Shaoting Zhang, and Dequan Wang.
\newblock Text-guided foundation model adaptation for pathological image classification.
\newblock In {\em International Conference on Medical Image Computing and Computer-Assisted Intervention}, pages 272--282. Springer, 2023.

\bibitem{zhu2023pointclip}
Xiangyang Zhu, Renrui Zhang, Bowei He, Ziyu Guo, Ziyao Zeng, Zipeng Qin, Shanghang Zhang, and Peng Gao.
\newblock Pointclip v2: Prompting clip and gpt for powerful 3d open-world learning.
\newblock In {\em Proceedings of the IEEE/CVF International Conference on Computer Vision}, pages 2639--2650, 2023.

\bibitem{kirillov2023segment}
Alexander Kirillov, Eric Mintun, Nikhila Ravi, Hanzi Mao, Chloe Rolland, Laura Gustafson, Tete Xiao, Spencer Whitehead, Alexander~C Berg, Wan-Yen Lo, et~al.
\newblock Segment anything.
\newblock {\em arXiv preprint arXiv:2304.02643}, 2023.

\bibitem{cheng2023sam}
Junlong Cheng, Jin Ye, Zhongying Deng, Jianpin Chen, Tianbin Li, Haoyu Wang, Yanzhou Su, Ziyan Huang, Jilong Chen, Lei Jiang, et~al.
\newblock Sam-med2d.
\newblock {\em arXiv preprint arXiv:2308.16184}, 2023.

\bibitem{wang2023sam}
Haoyu Wang, Sizheng Guo, Jin Ye, Zhongying Deng, Junlong Cheng, Tianbin Li, Jianpin Chen, Yanzhou Su, Ziyan Huang, Yiqing Shen, et~al.
\newblock Sam-med3d.
\newblock {\em arXiv preprint arXiv:2310.15161}, 2023.

\bibitem{du2023segvol}
Yuxin Du, Fan Bai, Tiejun Huang, and Bo~Zhao.
\newblock Segvol: Universal and interactive volumetric medical image segmentation.
\newblock {\em arXiv preprint arXiv:2311.13385}, 2023.

\bibitem{zhou2023foundation}
Yukun Zhou, Mark~A Chia, Siegfried~K Wagner, Murat~S Ayhan, Dominic~J Williamson, Robbert~R Struyven, Timing Liu, Moucheng Xu, Mateo~G Lozano, Peter Woodward-Court, et~al.
\newblock A foundation model for generalizable disease detection from retinal images.
\newblock {\em Nature}, 622(7981):156--163, 2023.

\bibitem{chowdhury2023can}
Pinaki~Nath Chowdhury, Ayan~Kumar Bhunia, Aneeshan Sain, Subhadeep Koley, Tao Xiang, and Yi-Zhe Song.
\newblock What can human sketches do for object detection?
\newblock In {\em Proceedings of the IEEE/CVF Conference on Computer Vision and Pattern Recognition}, pages 15083--15094, 2023.

\bibitem{zhang2024fm}
Dongmei Zhang, Chang Li, Renrui Zhang, Shenghao Xie, Wei Xue, Xiaodong Xie, and Shanghang Zhang.
\newblock Fm-ov3d: Foundation model-based cross-modal knowledge blending for open-vocabulary 3d detection.
\newblock In {\em Proceedings of the AAAI Conference on Artificial Intelligence}, volume~38, pages 16723--16731, 2024.

\bibitem{mazurowski2023segment}
Maciej~A Mazurowski, Haoyu Dong, Hanxue Gu, Jichen Yang, Nicholas Konz, and Yixin Zhang.
\newblock Segment anything model for medical image analysis: an experimental study.
\newblock {\em Medical Image Analysis}, 89:102918, 2023.

\bibitem{silva2023transductive}
Julio Silva-Rodr{\'\i}guez, Jose Dolz, and Ismail~Ben Ayed.
\newblock Transductive few-shot adapters for medical image segmentation.
\newblock {\em arXiv preprint arXiv:2303.17051}, 2023.

\bibitem{zhang2023customized}
Kaidong Zhang and Dong Liu.
\newblock Customized segment anything model for medical image segmentation.
\newblock {\em arXiv preprint arXiv:2304.13785}, 2023.

\bibitem{ma2024segment}
Jun Ma, Yuting He, Feifei Li, Lin Han, Chenyu You, and Bo~Wang.
\newblock Segment anything in medical images.
\newblock {\em Nature Communications}, 15(1):654, 2024.

\bibitem{dutt2023fairtune}
Raman Dutt, Ondrej Bohdal, Sotirios~A Tsaftaris, and Timothy Hospedales.
\newblock Fairtune: Optimizing parameter efficient fine tuning for fairness in medical image analysis.
\newblock {\em arXiv preprint arXiv:2310.05055}, 2023.

\bibitem{wu2023can}
Chaoyi Wu, Jiayu Lei, Qiaoyu Zheng, Weike Zhao, Weixiong Lin, Xiaoman Zhang, Xiao Zhou, Ziheng Zhao, Ya~Zhang, Yanfeng Wang, et~al.
\newblock Can gpt-4v (ision) serve medical applications? case studies on gpt-4v for multimodal medical diagnosis.
\newblock {\em arXiv preprint arXiv:2310.09909}, 2023.

\bibitem{ding2023parameter}
Ning Ding, Yujia Qin, Guang Yang, Fuchao Wei, Zonghan Yang, Yusheng Su, Shengding Hu, Yulin Chen, Chi-Min Chan, Weize Chen, et~al.
\newblock Parameter-efficient fine-tuning of large-scale pre-trained language models.
\newblock {\em Nature Machine Intelligence}, 5(3):220--235, 2023.

\bibitem{sagheer2020review}
Sameera V~Mohd Sagheer and Sudhish~N George.
\newblock A review on medical image denoising algorithms.
\newblock {\em Biomedical signal processing and control}, 61:102036, 2020.

\bibitem{willemink2020preparing}
Martin~J Willemink, Wojciech~A Koszek, Cailin Hardell, Jie Wu, Dominik Fleischmann, Hugh Harvey, Les~R Folio, Ronald~M Summers, Daniel~L Rubin, and Matthew~P Lungren.
\newblock Preparing medical imaging data for machine learning.
\newblock {\em Radiology}, 295(1):4--15, 2020.

\bibitem{kaissis2020secure}
Georgios~A Kaissis, Marcus~R Makowski, Daniel R{\"u}ckert, and Rickmer~F Braren.
\newblock Secure, privacy-preserving and federated machine learning in medical imaging.
\newblock {\em Nature Machine Intelligence}, 2(6):305--311, 2020.

\bibitem{chen2023dynamic}
Yuanyuan Chen, Xiaoqing Guo, Yongsheng Pan, Yong Xia, and Yixuan Yuan.
\newblock Dynamic feature splicing for few-shot rare disease diagnosis.
\newblock {\em Medical Image Analysis}, 90:102959, 2023.

\bibitem{dutt2024parameter}
Raman Dutt, Linus Ericsson, Pedro Sanchez, Sotirios~A Tsaftaris, and Timothy Hospedales.
\newblock Parameter-efficient fine-tuning for medical image analysis: The missed opportunity.
\newblock In {\em Medical Imaging with Deep Learning}, 2024.

\bibitem{wang2023medfmc}
Dequan Wang, Xiaosong Wang, Lilong Wang, Mengzhang Li, Qian Da, Xiaoqiang Liu, Xiangyu Gao, Jun Shen, Junjun He, Tian Shen, et~al.
\newblock A real-world dataset and benchmark for foundation model adaptation in medical image classification.
\newblock {\em Scientific Data}, 10(1):574, 2023.

\bibitem{jia2022visual}
Menglin Jia, Luming Tang, Bor-Chun Chen, Claire Cardie, Serge Belongie, Bharath Hariharan, and Ser-Nam Lim.
\newblock Visual prompt tuning.
\newblock In {\em European Conference on Computer Vision}, pages 709--727, 2022.

\bibitem{bahng2022exploring}
Hyojin Bahng, Ali Jahanian, Swami Sankaranarayanan, and Phillip Isola.
\newblock Exploring visual prompts for adapting large-scale models.
\newblock {\em arXiv preprint arXiv:2203.17274}, 2022.

\bibitem{wang2024universality}
Yihan Wang, Jatin Chauhan, Wei Wang, and Cho-Jui Hsieh.
\newblock Universality and limitations of prompt tuning.
\newblock {\em Advances in Neural Information Processing Systems}, 36, 2024.

\bibitem{hu2021lora}
Edward~J. Hu, Yelong Shen, Phillip Wallis, Zeyuan Allen-Zhu, Yuanzhi Li, Shean Wang, Lu~Wang, and Weizhu Chen.
\newblock Lora: Low-rank adaptation of large language models, 2021.

\bibitem{huang2024fpt}
Yijin Huang, Pujin Cheng, Roger Tam, and Xiaoying Tang.
\newblock Fpt: Fine-grained prompt tuning for parameter and memory efficient fine tuning in high-resolution medical image classification.
\newblock {\em arXiv preprint arXiv:2403.07576}, 2024.

\bibitem{zheng2024exploring}
Fudan Zheng, Jindong Cao, Weijiang Yu, Zhiguang Chen, Nong Xiao, and Yutong Lu.
\newblock Exploring low-resource medical image classification with weakly supervised prompt learning.
\newblock {\em Pattern Recognition}, 149:110250, 2024.

\bibitem{zhang2023understanding}
Chao Zhang, Xinyu Chen, Wensheng Li, Lixue Liu, Wei Wu, and Dacheng Tao.
\newblock Understanding deep neural networks via linear separability of hidden layers.
\newblock {\em arXiv preprint arXiv:2307.13962}, 2023.

\bibitem{he2023dvpt}
Along He, Kai Wang, Zhihong Wang, Tao Li, and Huazhu Fu.
\newblock Dvpt: Dynamic visual prompt tuning of large pre-trained models for medical image analysis.
\newblock {\em arXiv preprint arXiv:2307.09787}, 2023.

\bibitem{liu2023explicit}
Weihuang Liu, Xi~Shen, Chi-Man Pun, and Xiaodong Cun.
\newblock Explicit visual prompting for low-level structure segmentations.
\newblock In {\em Proceedings of the IEEE/CVF Conference on Computer Vision and Pattern Recognition}, pages 19434--19445, 2023.

\bibitem{zhou2022learning}
Kaiyang Zhou, Jingkang Yang, Chen~Change Loy, and Ziwei Liu.
\newblock Learning to prompt for vision-language models.
\newblock {\em International Journal of Computer Vision}, 130(9):2337--2348, 2022.

\bibitem{zhou2022conditional}
Kaiyang Zhou, Jingkang Yang, Chen~Change Loy, and Ziwei Liu.
\newblock Conditional prompt learning for vision-language models.
\newblock In {\em Proceedings of the IEEE/CVF conference on computer vision and pattern recognition}, pages 16816--16825, 2022.

\bibitem{yao2023visual}
Hantao Yao, Rui Zhang, and Changsheng Xu.
\newblock Visual-language prompt tuning with knowledge-guided context optimization.
\newblock In {\em Proceedings of the IEEE/CVF Conference on Computer Vision and Pattern Recognition}, pages 6757--6767, 2023.

\bibitem{chen2022adaptformer}
Shoufa Chen, GE~Chongjian, Zhan Tong, Jiangliu Wang, Yibing Song, Jue Wang, and Ping Luo.
\newblock Adaptformer: Adapting vision transformers for scalable visual recognition.
\newblock In {\em Advances in Neural Information Processing Systems}, 2022.

\bibitem{zhang2022tip}
Renrui Zhang, Wei Zhang, Rongyao Fang, Peng Gao, Kunchang Li, Jifeng Dai, Yu~Qiao, and Hongsheng Li.
\newblock Tip-adapter: Training-free adaption of clip for few-shot classification.
\newblock In {\em European conference on computer vision}, pages 493--510. Springer, 2022.

\bibitem{yu2023task}
Tao Yu, Zhihe Lu, Xin Jin, Zhibo Chen, and Xinchao Wang.
\newblock Task residual for tuning vision-language models.
\newblock In {\em Proceedings of the IEEE/CVF Conference on Computer Vision and Pattern Recognition}, pages 10899--10909, 2023.

\bibitem{gao2024clip}
Peng Gao, Shijie Geng, Renrui Zhang, Teli Ma, Rongyao Fang, Yongfeng Zhang, Hongsheng Li, and Yu~Qiao.
\newblock Clip-adapter: Better vision-language models with feature adapters.
\newblock {\em International Journal of Computer Vision}, 132(2):581--595, 2024.

\bibitem{zhao2024sct}
Henry~Hengyuan Zhao, Pichao Wang, Yuyang Zhao, Hao Luo, Fan Wang, and Mike~Zheng Shou.
\newblock Sct: A simple baseline for parameter-efficient fine-tuning via salient channels.
\newblock {\em International Journal of Computer Vision}, 132(3):731--749, 2024.

\bibitem{wu2023medical}
Junde Wu, Rao Fu, Huihui Fang, Yuanpei Liu, Zhaowei Wang, Yanwu Xu, Yueming Jin, and Tal Arbel.
\newblock Medical sam adapter: Adapting segment anything model for medical image segmentation.
\newblock {\em arXiv preprint arXiv:2304.12620}, 2023.

\bibitem{song2023vppt}
Zhao Song, Ke~Yang, Naiyang Guan, Junjie Zhu, Peng Qiao, and Qingyong Hu.
\newblock Vppt: Visual pre-trained prompt tuning framework for few-shot image classification.
\newblock In {\em ICASSP 2023-2023 IEEE International Conference on Acoustics, Speech and Signal Processing (ICASSP)}, pages 1--5. IEEE, 2023.

\bibitem{yaras2023law}
Can Yaras, Peng Wang, Wei Hu, Zhihui Zhu, Laura Balzano, and Qing Qu.
\newblock The law of parsimony in gradient descent for learning deep linear networks.
\newblock {\em arXiv preprint arXiv:2306.01154}, 2023.

\bibitem{dosovitskiy2020image}
Alexey Dosovitskiy, Lucas Beyer, Alexander Kolesnikov, Dirk Weissenborn, Xiaohua Zhai, Thomas Unterthiner, Mostafa Dehghani, Matthias Minderer, Georg Heigold, Sylvain Gelly, et~al.
\newblock An image is worth 16x16 words: Transformers for image recognition at scale.
\newblock {\em arXiv preprint arXiv:2010.11929}, 2020.

\bibitem{yosinski2014transferable}
Jason Yosinski, Jeff Clune, Yoshua Bengio, and Hod Lipson.
\newblock How transferable are features in deep neural networks?
\newblock {\em Advances in neural information processing systems}, 27, 2014.

\bibitem{codella2019skin}
Noel Codella, Veronica Rotemberg, Philipp Tschandl, M~Emre Celebi, Stephen Dusza, David Gutman, Brian Helba, Aadi Kalloo, Konstantinos Liopyris, Michael Marchetti, et~al.
\newblock Skin lesion analysis toward melanoma detection 2018: A challenge hosted by the international skin imaging collaboration (isic).
\newblock {\em arXiv preprint arXiv:1902.03368}, 2019.

\bibitem{grand}
Medfmc challenge.
\newblock https://medfm2023.grand-challenge.org/medfm2023.

\bibitem{Mmclassification}
Mmclassification.
\newblock https://github.com/open-mmlab/mmclassification.

\bibitem{Bahng_Jahanian_Sankaranarayanan_Isola_2022}
Hyojin Bahng, Ali Jahanian, Swami Sankaranarayanan, and Phillip Isola.
\newblock Exploring visual prompts for adapting large-scale models.
\newblock Mar 2022.

\bibitem{jha2020kvasir}
Debesh Jha, Pia~H Smedsrud, Michael~A Riegler, P{\aa}l Halvorsen, Thomas De~Lange, Dag Johansen, and H{\aa}vard~D Johansen.
\newblock Kvasir-seg: A segmented polyp dataset.
\newblock In {\em MultiMedia Modeling: 26th International Conference, MMM 2020, Daejeon, South Korea, January 5--8, 2020, Proceedings, Part II 26}, pages 451--462. Springer, 2020.

\bibitem{jha2020medico}
Debesh Jha, Steven~A Hicks, Krister Emanuelsen, H{\aa}vard Johansen, Dag Johansen, Thomas de~Lange, Michael~A Riegler, and P{\aa}l Halvorsen.
\newblock Medico multimedia task at mediaeval 2020: Automatic polyp segmentation.
\newblock {\em arXiv preprint arXiv:2012.15244}, 2020.

\bibitem{MMSegmentation}
Mmsegmentation.
\newblock https://github.com/open-mmlab/mmsegmentation.

\bibitem{zheng2021rethinking}
Sixiao Zheng, Jiachen Lu, Hengshuang Zhao, Xiatian Zhu, Zekun Luo, Yabiao Wang, Yanwei Fu, Jianfeng Feng, Tao Xiang, Philip~HS Torr, et~al.
\newblock Rethinking semantic segmentation from a sequence-to-sequence perspective with transformers.
\newblock In {\em Proceedings of the IEEE/CVF conference on computer vision and pattern recognition}, pages 6881--6890, 2021.

\bibitem{ronneberger2015u}
Olaf Ronneberger, Philipp Fischer, and Thomas Brox.
\newblock U-net: Convolutional networks for biomedical image segmentation.
\newblock In {\em Medical Image Computing and Computer-Assisted Intervention--MICCAI 2015: 18th International Conference, Munich, Germany, October 5-9, 2015, Proceedings, Part III 18}, pages 234--241. Springer, 2015.

\end{thebibliography}
\bibliographystyle{unsrt}

\clearpage
\appendix
\onecolumn
\section{More Results and Details}

\subsection{Details of Implementation}
\subsubsection{MedFMC Classification Experiments}
Our experimental setups are consistent with MedFMC Challenge~\cite{grand} and MedFMC dataset~\cite{wang2023medfmc}. Although Wang et al.~\cite{wang2023medfmc} conducted extensive experiments, they did not provide results on mainstream architectures such as ViT. We run the fine-tuning experiments using ViT(ViT-Base/16). The ViT model is optimized with an initial learning rate of 6e-4. These classification models are trained on a single NVIDIA A100 for 20 epochs with a batch size of 4, employing the MMClassification framework~\cite{Mmclassification}.

\paragraph{\textbf{EPT}} We utilize the validation set of each dataset to determine the optimal prompt length $d_p$. The prompt length represents the exclusive hyperparameter of the EPT approach that we adjust. For Transformer-based models, we consider the range of values for prompt length as [5, 10, 20, 50, 100, 200, 400, 600]. For the ViT architecture, we set the prompted layers to 'DEEP', which means prompts are used in all layers. 

Note: For ViT-base/16, the number of embedding patches is 196. Embedding dimensions for each patch are 768. Consequently, to facilitate comparison with VPT, we multiply the relative prompt length by 196/768. Each prompt is randomly initialized using a normal initialization scheme. We adhere to the original design choices of the backbone architecture.

\paragraph{\textbf{VPT}} For the settings in VPT, we follow Jia et al.~\cite{jia2022visual}. We use the validation set of each dataset to find the best prompt length $n_p$. The prompt length range for Transformer backbones spans [1, 5, 10, 50, 100, 150]. For the ViT architecture, we set the prompted layers to 'DEEP', which means prompts are used in all layers.

\paragraph{\textbf{VP}} For the settings in VP~\cite{Bahng_Jahanian_Sankaranarayanan_Isola_2022}, we use the validation set of each dataset to find the best prompt length $n_p$. Since the VP method is not our primary comparative approach, we only conducted experiments on the ViT (Vision Transformer) architecture. The range of prompt length is [1, 5, 10, 50, 100, 150]. We set the prompted layers to 'DEEP', which means prompts are used in all layers.

\paragraph{\textbf{Adapters}}
Adapters ~\cite{chen2022adaptformer} insert extra lightweight modules inside each Transformer layer. One adapter module generally consists of a linear down-projection (with a reduction rate r), followed by a nonlinear activation function, and a linear up-projection, together with a residual connection. Chen et al.~\cite{chen2022adaptformer} explored the insertion method of adapters. Therefore we also use this setup in our own implementation. We choose the reduction rate r in [1, 4, 8, 32, 64].

\paragraph{\textbf{LoRA}} According to the findings of LoRA~\cite{hu2021lora}, we applied the LoRA structure to the matrices $W_q$ and $W_v$, and we chose the reduction rate $r$ from the options [1, 4, 8, 32, 64] to select the best for each task.

\begin{table*}[!ht]
\captionsetup{singlelinecheck=off, justification=raggedright,labelsep=period}
\caption{Fine-tuning time of 10 shot learning on Colon dataset with different methods.}
\centering
\begin{tabular}{cccc}
\hline Method & Convergence & Time \\
\hline  Full & 20 epochs & 5min02s \\
\hline Linear & 20 epochs & 2min06s \\
Partial-1 & 20 epochs & 2min16s  \\
MLP-3 & 20 epochs & 2min36s   \\
\hline LoRA & 20 epochs & 2min09s \\
Bias & 20 epochs & 2min08s \\
Adapter & 20 epochs & 2min07s  \\
\hline VP & 20 epochs & 2min08s   \\
VPT & 20 epochs & 2min42s  \\
EPT & 20 epochs & 2min08s \\
\hline
\end{tabular}
\label{tab:time}
\end{table*}
\paragraph{\textbf{Fine-tuning time}} We selected the Colon-10shot task to demonstrate the time and number of epochs needed for fine-tuning, as it is the most computationally demanding task among all. For few-shot fine-tuning tasks, as evident from Tab \ref{tab:time}, Moreover, the entire process can generally be completed within 2-5 minutes. The full fine-tuning method takes the longest computation time, followed by VPT and MLP-3. Our EPT method partially mitigates the computational time drawback associated with prompt-based methods and achieves a computational time close to that of the Adapter method in few-shot scenarios.

\subsubsection{\zdiff{ISIC 2018 Classification Experiments}}
\zdiff{The ISIC 2018 dataset~\cite{codella2019skin}  contained 10,015 training and 1512 test images for a total of 11,527 images. ISIC 2018 provided the ground-truth data, consisting of seven classes, melanoma,melanocytic nevus, basal cell carcinoma, intraepithelial carcinoma,benign keratosis,dermatofibroma and vascular lesion. We strive to maintain consistency with the MedFMC classification tasks. The task is also formulated as N-Way K-shot classification task, and K is set to 1,5,10. 1-shot refers to every N images that sample from each class, rather than one image.}

\subsubsection{Segmentation Experiments}
We select kvasir-SEG ~\cite{jha2020kvasir} for polyp segmentation task. We follow the settings in Medico automatic polyp segmentation· task at mediaeval 2020 ~\cite{jha2020medico} with a train-valid ratio of 880:120. 

For a more meaningful comparison, we have employed SETR as our segmentation framework to ensure consistency with VPT. We have utilized the same hyperparameters as in MMSegmentation~\cite{MMSegmentation}. However, we have opted to use ViT-Base/16 pretrained on ImageNet-21k as our backbone.

\begin{table*}[!ht]
\captionsetup{singlelinecheck=off, justification=raggedright,labelsep=period}
    \caption{Semantic Segmentation: kvasir-SEG~\cite{jha2020kvasir} validation results with SETR~\cite{zheng2021rethinking} on ViT-B/16. The best mIoU scores among all methods but Full are bolded. Results of fully fine-tuning a FCN-UNet~\cite{ronneberger2015u} are included.}
    \centering
\begin{tabular}{cc|ccccc|c}
\hline Backbone & \multicolumn{6}{c}{ ViT-B/16 } & \multicolumn{1}{c}{ FCN-UN } \\
\cline { 2 - 8 } Method & Full  & Head & Bias & Adapter & VPT & EPT & Full \\
\hline mIoU & 55.97 & 53.62 & 53.21 & $\mathbf{56.77}$ & 52.99 & 53.57 & 70.77 \\
mDice & 69.00 & 67.18& 65.88 & $\mathbf{69.41}$ & 64.73 & 66.21 & 81.58 \\
Tunable params (M) & 87.76 & 1.33& 1.43 & 2.52 & 2.25 & 1.94 & 28.99 \\

\hline
\end{tabular}
    \label{tab:seg}
\end{table*}
\paragraph{\textbf{Apply EPT to segmentation}} We explored the effectiveness of fine-tuning algorithms on the medical segmentation dataset, kvasir-SEG~\cite{jha2020kvasir}. We employed the same segmentation algorithm, SETR-PUP~\cite{zheng2021rethinking}, as used in VPT. For the ViT backbone, we utilized the ViT-Base/16 model pre-trained on ImageNet-21k. However, we did not use a pre-trained segmentation head. The standard practice (Full) involves fully fine-tuning the pre-trained backbone alongside the ConvNet head. For comparison, three additional protocols are included: Head, VPT, and adapter. Additionally, we used full fine-tuning FCN-UN as a comparative benchmark. The results are shown in Tab \ref{tab:seg}. A major difference is that our EPT method lags behind in the segmentation task, with a lower mIoU compared to the Head method, but outperforms VPT. The Adapter method achieves the best results in the segmentation task, surpassing the results of full fine-tuning. The reasons behind these results are still unclear, and further exploration is needed for our EPT in segmentation experiments.

\subsection{Proofs}
\textbf{Proof for Lemma \ref{lemma}}

Proof. For any $i\in [n_k]$, we have $\mathbf{x}_{k, i} \geqslant 0$. Therefore $\bar{\mathbf{x}}_k \geqslant 0$ and for $\mathbf{x}_{k, i}$, $\bar{\mathbf{x}}_k$, we use $\theta\left(\mathbf{x}_{k, i}, \bar{\mathbf{x}}_k\right)$ to represent the angle between two vectors $\mathbf{x}_{k, i}$ and $\bar{\mathbf{x}}_k$.
Thus, we have $\theta\left(\mathbf{x}_{k, i}, \bar{\mathbf{x}}_k\right) \leqslant \frac{\pi}{2}$.

\begin{equation}
  \left\|c_k \mathbf{x}_{k,i}-c_k \bar{\mathbf{x}}_k\right\|_2=\left(\left\|c_k \mathbf{x}_{k, i}\right\|_2^2+\left\|c_k \bar{\mathbf{x}}_k\right\|_2^2-2\left\|c_k \mathbf{x}_{k, i}\right\|_2|| c_k \bar{\mathbf{x}}_k \|\cos \left(\theta\left(\mathbf{x}_{k, i}, \bar{\mathbf{x}}_k\right)\right)\right)^{\frac{1}{2}}
  \label{eq:1}
\end{equation}

\begin{equation}
\left\|c_{k, i} \mathbf{x}_{k, i}-c_k \bar{\mathbf{x}_k}\right\|_2=\left(\left\|c_{k, i} \mathbf{x}_{k, i}\right\|_2^2+\left\|c_k \bar{\mathbf{x}}_k\right\|_2^2-2\left\|c_{k, i} \mathbf{x}_{k, i}\right\|_2\left\|c_k \bar{\mathbf{x}}_k\right\|_2 \cos \left(\theta\left(\mathbf{x}_{k, i}, \bar{\mathbf{x}}_k\right)\right)\right)^{\frac{1}{2}}
  \label{eq:2}
\end{equation}

Consider two cases:(1)$\left\|\mathbf{x}_{k, i}\right\|_2 \leqslant\left\|\bar{\mathbf{x}}_k\right\|_2$;(2)$\left\|\mathbf{x}_{k,i}\right\|_2 \geqslant \left\|\bar{\mathbf{x}}_k \right\|_2$.

For $\left\|\mathbf{x}_{k, i}\right\|_2 \leqslant\left\|\bar{\mathbf{x}}_k\right\|_2$, we have $\left\|c_k \mathbf{x}_{k, i}\right\|_2\leqslant \left\|c_{k, i} \mathbf{x}_{k, i}\right\|_2\leqslant \left\|c_k \bar{\mathbf{x}}_k\right\|_2$. According to \eqref{eq:1} and \eqref{eq:2}, we have $\left\|c_{k, i} \mathbf{x}_{k, i}-c_k \bar{\mathbf{x}}_k\right\|_2\leqslant c_k\left\|\mathbf{x}_{k, i}-\bar{\mathbf{x}}_k\right\|_2$.

And for $\left\|\mathbf{x}_{k,i}\right\|_2 \geqslant \| \bar{\mathbf{x}}_k \|_2$, we have $\left\|c_k \mathbf{x}_{k, i}\right\|_2\geqslant \left\|c_{k, i} \mathbf{x}_{k, i}\right\|_2\geqslant \left\|c_k \bar{\mathbf{x}}_k\right\|_2$. According to \eqref{eq:1} and \eqref{eq:2}, we have $\left\|c_{k, i} \mathbf{x}_{k, i}-c_k \bar{\mathbf{x}}_k\right\|_2\leqslant c_k\left\|\mathbf{x}_{k, i}-\bar{\mathbf{x}}_k\right\|_2$.

So we can conclude that $\Sigma_\mathbf{W}^{\prime} \leqslant c_k^2 \Sigma_\mathbf{W}$.

\textbf{Proof for Proposition \ref{prop}}

Proof. According to the assumptions of Proposition \ref{prop}, assume $d_p=1$ and $n=2$. Let $\mathbf{\sigma}_1=\left[\sigma_{11}, \sigma_{12}\right]$ and $\mathbf{\sigma}_2=\left[\sigma_{21}, \sigma_{22}\right]$ represent $\sigma\left(\mathbf{u}_1\right)$ and $\sigma\left(\mathbf{u}_2\right)$ respectively. Also, let $\mathbf{\sigma}_{p,1}=\left[\sigma_{p,11}, \sigma_{p,12}\right]$ and $\mathbf{\sigma}_{p,2}=\left[\sigma_{p,21}, \sigma_{p,22}\right]$ represent $\sigma\left(\mathbf{u}_{p, 1}\right)_{d_p:}$ and $\sigma\left(\mathbf{u}_{p, 2}\right)_{d_p:}$ respectively.


Then $\sigma_{11}+\sigma_{12}=\sigma_{21}+\sigma_{22}=1$ and $\mathbf{\sigma}_1>0, \mathbf{\sigma}_2>0$. Without loss of generality, assume $\left\|\mathbf{\sigma}_1\right\|_2 \leqslant\left\|\mathbf{\sigma}_2\right\|_2$. It can be inferred that $\left\|\sigma_{11}-\sigma_{12}\right\|_2 \leqslant\left\|\sigma_{21}-\sigma_{22}\right\|_2$.

Let $\mathbf{\sigma}_1=\left[\frac{1}{1+e^{z_1}}, \frac{e^{z_1}}{1+e^{z_1}}\right]$ and $\mathbf{\sigma}_2=\left[\frac{1}{1+e^{z_2}}, \frac{e^{z_2}}{1+e^{z_2}}\right]$.
Then we have $z_2>z_1$. 

For $\mathbf{\sigma}_{p,1}$ and $\mathbf{\sigma}_{p,2}$, let $\mathbf{u}_{p,1} =\left[\frac{1}{1+e^{z_1}+e^{z_1 p}}, \frac{e^{z_1}}{1+e^{z_1}+e^{z_1 p}}\right]$ and $\mathbf{u}_{p,2} =\left[\frac{1}{1+e^{z_2}+e^{z_2 p}}, \frac{e^{z_2}}{1+e^{z_2}+e^{z_2 p}}\right]$.

Therefore, for $\mathbf{\sigma}_{p, 1}=c_1 \mathbf{\sigma}_1$ and $\mathbf{\sigma}_{p, 2}=c_2 \sigma_2$, we can know that $c_1=\frac{1+e^{z_1}}{1+e^{z_1}+e^{z_1 p}}$ and $c_2=\frac{1+e^{z_2}}{1+e^{z_2}+e^{z_2 p}}$.

To prove: $c_1\geqslant c_2$, it is equivalent to proving: $\frac{e^{z_1 p}}{1+e^{z_1}+e^{z_1 p}}\leqslant \frac{e^{z_2 p}}{1+e^{z_2}+e^{z_2 p}}$, given that $z_2\geqslant z_1$. Therefore, we can conclude that $p=1$ satisfies the condition. Thus, $c_1=\frac{\left\|\sigma\left(\mathbf{u}_{p,1}\right)_{d_p:}\right\|_2}{\left\|\sigma\left(\mathbf{u}_1\right)\right\|_2} \geqslant c_2=\frac{\left\|\sigma\left(\mathbf{u}_{p,2}\right)_{d_p:}\right\|_2}{\left\|\sigma\left(\mathbf{u}_2\right)\right\|_2}$.

\end{document}